# Partial Sum Minimization of Singular Values in Robust PCA: Algorithm and Applications

Tae-Hyun Oh *Student Member, IEEE,* Yu-Wing Tai *Senior Member, IEEE,* Jean-Charles Bazin *Member, IEEE,* Hyeongwoo Kim *Student Member, IEEE,* and In So Kweon *Member, IEEE*

**Abstract**—Robust Principal Component Analysis (RPCA) via rank minimization is a powerful tool for recovering underlying low-rank structure of clean data corrupted with sparse noise/outliers. In many low-level vision problems, not only it is known that the underlying structure of clean data is low-rank, but the exact rank of clean data is also known. Yet, when applying conventional rank minimization for those problems, the objective function is formulated in a way that does not fully utilize a priori target rank information about the problems. This observation motivates us to investigate whether there is a better alternative solution when using rank minimization. In this paper, instead of minimizing the nuclear norm, we propose to minimize the partial sum of singular values, which implicitly encourages the target rank constraint. Our experimental analyses show that, when the number of samples is deficient, our approach leads to a higher success rate than conventional rank minimization, while the solutions obtained by the two approaches are almost identical when the number of samples is more than sufficient. We apply our approach to various low-level vision problems, e.g. high dynamic range imaging, motion edge detection, photometric stereo, image alignment and recovery, and show that our results outperform those obtained by the conventional nuclear norm rank minimization method.

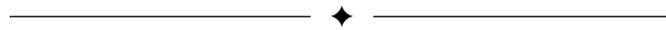

## 1 INTRODUCTION

Various low-level vision applications, including High Dynamic Range (HDR) [35], [36], photometric stereo [3], [23], batch image alignment [38] and factorization-based structure from motion [5], [41], can be formulated as a low-rank matrix recovery problem. Low-rank matrix approximation methods, such as Principal Component Analysis (PCA) [29] and matrix factorization [7], [18], [40], [48] are widely used to find the best approximation of an underlying low-rank structure of data. However many of these approaches are error-prone due to the presence of outliers. To recover the low-rank matrix while rejecting outliers, a rank minimization based Robust Principal Component Analysis (RPCA) [9] has been proposed and gained much interest in computer vision [28], [38], [45], [47].

RPCA [9] aims to recover a low-rank matrix $\mathbf{A} \in \mathbb{R}^{m \times n}$ from corrupted observations $\mathbf{O} = \mathbf{A} + \mathbf{E}$, where $\mathbf{E} \in \mathbb{R}^{m \times n}$ represents errors with arbitrary magnitude and distribution. The rank minimization approach [9], [10], [39], [43] assumes $\mathbf{E}$ is sparse and formulates the problem as:

$$\min_{\mathbf{A},\mathbf{E}} \operatorname{rank}(\mathbf{A}) + \lambda\|\mathbf{E}\|_0, \quad \text{s.t. } \mathbf{O} = \mathbf{A} + \mathbf{E}, \quad (1)$$

where $\|\cdot\|_0$ denotes the $l^0$-norm, and $\lambda$ is the relative weight between the two terms. Unfortunately, solving the problem

- T.-H. Oh, Y.-W. Tai, H. Kim and I.S. Kweon (corresponding author) are with the Department of Electrical Engineering, KAIST, Daejeon, Republic of Korea.
  E-mail: thoh.kaist.ac.kr@gmail.com, yuwing@gmail.com, hyeongwoo.kim@kaist.ac.kr, iskweon77@kaist.ac.kr
- J.-C. Bazin is with the Department of Computer Science, ETH Zurich, Switzerland.
  E-mail: jean-charles.bazin@inf.ethz.ch

in Eq. (1) is an NP-hard problem. The rank minimization approach instead solves an approximated problem by convex surrogate as:

$$\min_{\mathbf{A},\mathbf{E}} \|\mathbf{A}\|_* + \lambda\|\mathbf{E}\|_1, \quad \text{s.t. } \mathbf{O} = \mathbf{A} + \mathbf{E}, \quad (2)$$

where $\|\mathbf{A}\|_* = \sum_i \sigma_i(\mathbf{A})$ is the nuclear norm of $\mathbf{A}$, $\sigma_i(\mathbf{A})$ represents the $i$-th singular value of $\mathbf{A}$ (sorted in decreasing order), and $\|\mathbf{E}\|_1$ is the $l^1$-norm of $\mathbf{E}$. Eq. (2) can be solved effectively by various methods [31], [44]. Wright *et al.* [43] and Candès *et al.* [9] proved that, under mild conditions, the unique solution of Eq. (2) *exactly* corresponds to the solution of the original NP-hard problem in Eq. (1). Yet, when the number of observations in $\mathbf{O}$ is very limited, experiments show that the converged solution includes some outliers as inliers and vice versa. Such limited number of observations is not uncommon in many computer vision problems due to practical reasons. For example, in HDR context, often only 2-4 differently exposed images are captured and photometric stereo requires only 3 images in theory. Moreover, we also observe that the converged solution can be degenerated. For instance, in the photometric stereo problem [45], the solution of $\mathbf{A}$ might have a rank lower than the theoretical rank of 3.

In this paper, based on the prior knowledge about the rank of $\mathbf{A}$, we propose an alternative objective function which minimizes the Partial Sum of Singular Values (abbreviated to PSSV) of $\mathbf{A}$:

$$\min_{\mathbf{A},\mathbf{E}} \|\mathbf{A}\|_{p=N} + \lambda\|\mathbf{E}\|_1, \quad \text{s.t. } \mathbf{O} = \mathbf{A} + \mathbf{E}, \quad (3)$$

where $\|\mathbf{A}\|_p = \sum_{i=p+1}^{\min(m,n)} \sigma_i(\mathbf{A})$ and $N$ is the target rank of $\mathbf{A}$ which can be derived from the problem definition, e.g. $N=1$ for background subtraction, $N=3$ for photometric stereo. Eq. (3) minimizes the rank of residual errors of $\mathbf{A}$

2against the target rank, instead of the nuclear norm. A major drawback of using the nuclear norm to approximate the $l^0$-norm of singular values is that the nuclear norm minimizes not only the rank of $\mathbf{A}$, but also the variance of $\mathbf{A}$ by simultaneously minimizing all the singular values of $\mathbf{A}$ including the singular values within the target rank $N$, i.e. $\sigma_{1 \leq i \leq N}$. Consequently, the low-rank structure of $\mathbf{A}$ may not be well approximated under the environment that does not follow the assumption of large number of inputs.

Although Eq. (3) is non-convex, we observe in our experiments that Eq. (3) encourages the resulting low-rank matrix to have a rank close to $N$ even with deficient observations. For example, when the singular values of $\mathbf{A}$ within the target rank are small, the nuclear norm can result in a rank deficient matrix $\mathbf{A}$, i.e. whose rank is smaller than the target rank. In contrast, because our work does not minimize the subspace variance of $\mathbf{A}$ within the target rank, we are not biased to the solution with smaller variance of $\mathbf{A}$. Thus, the low-rank matrix $\mathbf{A}$ can be more accurately estimated. Further analyses and discussions about this claim are provided in the later sections of our paper.

In contrast to matrix factorization methods where the a priori rank constraint is enforced as a hard constraint via matrix projection, we encourage the rank constraint as a soft constraint and propose the Partial Singular Value Thresholding (PSVT) to solve our partial sum singular value objective function. As analyzed in our study, the PSVT operator encourages the result $\mathbf{A}$ to meet the target rank even when all the singular values are small.

This work is the extension of our previous conference paper [33]. We empirically study the proposed objective function in many low-level vision problems, e.g. HDR imaging, motion boundary detection, photometric stereo, image alignment, and image recovery where the theoretical rank of $\mathbf{A}$ is known and the number of observations is limited (except the image recovery application). Our experimental analyses show that our formulation, described in Eq. (3), converges to a solution more robust to outliers than the solution obtained by the objective function in Eq. (2) in rank minimization, when the number of observations is limited. Empirically, we also find that the solutions of Eq. (2) (nuclear norm) and Eq. (3) (our PSSV) are almost identical when more than a sufficient number of samples is observed.

In short summary, our contributions are as follows:

- We present a partial sum objective function and its corresponding minimization method for RPCA.
- We empirically study the partial sum objective function and claim that it outperforms the nuclear norm rank minimization when the number of observed samples is very limited.
- We present the convergence property of the proposed algorithm to minimize the proposed partial sum objective function consisting of PSSV and sparse term, and provide its proof.
- We apply our technique on various low-level vision problems and demonstrate superior results over previous works.

## 2 RELATED WORKS

In this section, we briefly review early works related to RPCA, then we discuss some recent advances in RPCA and its applications in computer vision. We will also review some recent matrix factorization based works for low-rank approximation. We invite readers to refer to Candès et al. [9] for a thoughtful review of RPCA.

In conventional PCA [29], the goodness-of-fit of data is evaluated by the $l^2$-norm which is very sensitive to outliers. Early works in RPCA tried to reduce the effects of outliers by random sampling [19], robust M-estimator [12], [13], or alternating minimization [30] to identify outliers or penalize data with large errors. These methods share some limitations: either they are sensitive to the choice of parameters or their algorithms are not scalable enough in running time.

Recent advances in RPCA showed that the heuristic nuclear norm solution [9], [39], [43] converges to a solution which is robust to sparse outliers. Candès et al. [9] proved that the original RPCA problem as in Eq. (1) can be solved by instead solving the convex relaxation version in Eq. (2), and it provides a unique and *exact* solution of Eq. (1) as long as $\mathbf{E}$ is sparse and random and the underlying $\text{rank}(\mathbf{A})$ is lower than a certain upper bound[1]. To solve Eq. (2), various methods have been proposed [31], [44]. Among them, Alternating Direction Method of Multipliers (ADMM, or also called inexact augmented Lagrange multiplier) [31] has shown to be computationally efficient. Also, Zhou et al. [49] and Agarwal et al. [1] proved that convex approximation by nuclear norm can still achieve bounded and stable results even under small noise measurements.

Besides the standard nuclear norm relaxation, some works study variants of the nuclear norm to enhance performance of rank minimization [11], [20], [24]. Chen et al. [11] and Gaiffas et al. [20] proposed an adaptive weighted nuclear norm. They suggested a non-trivial update scheme to update the adaptive weighted nuclear norm and claimed to achieve higher accuracy in low-rank matrix approximation in comparison with the standard nuclear norm. Hu et al. [24] proposed the truncated nuclear norm (TNN) for the matrix completion problem which shares a similar objective function with our PSSV objective function. Since the TNN is non-convex (which is not easy to directly solve), they aim to avoid direct minimization by locally approximating TNN as $\min_{X,W} \|X\|_* - \text{Tr}(A_l W B_l^\top)$, s.t. $X = W$ by alternatively minimizing $A_l, B_l, X$ and $W$ based on the singular value thresholding (SVT) operator [8]. This alternating scheme requires outer iterations and additional SVD computations. Instead of utilizing this alternating scheme, we propose the PSVT operator to directly minimize the partial sum of singular value term. Although our objective function is also non-convex, our proposed PSVT produces the closed-form solution to the sum of the PSSV and proximity term. In that sense, every sub-problem of our problem has a closed-form solution. Thus, our optimization problem can be solved efficiently. Moreover, while the approach of Hu et al. is dedicated only to matrix completion, we show that our work can be successfully applied for several computer

---

1. The bound depends on the matrix size.

vision tasks spanning from image alignment to photometric stereo and HDR imaging.

Another branch of low-rank framework is based on Matrix Factorization (MF). Several robust MF methods based on $l^1$-norm have been suggested [7], [18], [48]. A benefit of matrix factorization approaches is that they can easily enforce the rank constraint by the explicit bilinear matrix form. The target rank constraint is enforced as a hard constraint via matrix reprojection or orthogonal procrustes. Cabral *et al.* [7] revisited the relationship between nuclear norm regularization and bilinear MF model [2], and proposed a rank continuation heuristic to avoid local minima. Compared with MFs, our target rank constraint is expressed as a soft constraint which provides flexibility to balance between the rank constraint and other constraints used in different computer vision problems.

The robustness and scalability of the rank minimization algorithm for RPCA [9], [31], [44] have inspired many applications in computer vision, such as background subtraction [9], image and video restoration [28], image alignment [38], regular texture analysis [47], and robust photometric stereo [45]. These applications are based on the observation that the underlying structures of clean data are linearly correlated, which forms a low-rank data matrix. The rank minimization proposed by Candès *et al.* [9] is general in the sense that it does not require to know a priori the rank of clean data. However, as briefly mentioned in the introduction, in some applications, the rank of clean data can be determined by the problem definition, and this motivates us to investigate how the prior rank information can be fully utilized in the context of RPCA.

The success of rank minimization based RPCA comes from the blessing of dimensionality of input matrix [16], [43], implying large amount of observations. However, when the number of observations is limited, which is common in practice, results from RPCA might be degenerated, e.g. correct samples might be considered as outliers and vice versa. As discussed in the introduction, this happens because the standard nuclear norm minimizes not only the rank of the matrix, but also the variance of data distribution of the matrix. To overcome this limitation, we introduce an alternative objective function that can efficiently deal with a deficient number of samples in the rank minimization problem. Our work can be considered as an addendum to the standard rank minimization approach when the target rank or the approximate target rank is known. The proposed alternative objective function can control the rank with a simple and efficient minimizer. We demonstrate the effectiveness of our proposed objective function through thoughtful experiments.

## 3 PARTIAL SUM MINIMIZATION BY THE PSVT OPERATOR

### 3.1 Derivation of Partial Sum of Singular Values

Our partial sum formulation in Eq. (3) is originated from the following objective function:

$$\arg\min_{\mathbf{A},\mathbf{E}} \ |\text{rank}(\mathbf{A}) - N| + \lambda \|\mathbf{E}\|_0, \ \text{s.t.} \ \mathbf{O} = \mathbf{A} + \mathbf{E}. \quad (4)$$

Eq. (4) aims to recover a low-rank matrix $\mathbf{A}$ close to the target rank $N$ and a sparse error matrix $\mathbf{E}$.

Since the above objective function is also an NP-hard problem, we relax it with an alternative representation in order to effectively deal with it. The relaxation is similar to the method presented by Candès *et al.* [9]. We should also properly interpret the target rank $N$. We relax it with a projection operator to enforce a rank-$N$ matrix in a matrix interpretation manner. From the relaxation, the PSSV objective function, which is the first term in Eq. (4), can be derived as follows:

$$\begin{aligned} \|\mathbf{A}\|_* - \|P_N(\mathbf{A})\|_* &= \left| \sum_{i=1}^{\min(m,n)} \sigma_i(\mathbf{A}) - \sum_{i=1}^{N} \sigma_i(\mathbf{A}) \right| \\ &= \sum_{i=N+1}^{\min(m,n)} \sigma_i(\mathbf{A}) = \|\mathbf{A}\|_{p=N}, \end{aligned} \quad (5)$$

where $\|\cdot\|_{p=N}$ denotes the PSSV with the target rank $N$, and $P_r(\cdot)$ is the matrix projection operator to rank-$r$ matrix defined as follows.

**Definition 1. [Projection operator]**

$$P_r(\mathbf{X}) = \mathbf{U}_{1:r}^\top \mathbf{X} \mathbf{V}_{1:r}, \quad (6)$$

where $\mathbf{U}_{1:r}$ and $\mathbf{V}_{1:r}$ are the matrices consisting of the singular vectors corresponding to the $r$ largest singular values of $\mathbf{X}$.

#### 3.1.1 From rank constraint to projection

Eq. (5) leads us to the PSSV objective function in Eq. (3). In this section, we show the relationship between the target rank $N$ and the projection operator in Eq. (5). We first introduce a rank representation.

**Lemma 1.** *Let $\mathbf{A} \in \mathbb{R}^{m \times n}$ and $\text{rank}(\mathbf{A}) \geq r$, then there exist matrices $\mathbf{C} \in \mathbb{R}^{r \times m}$ and $\mathbf{B} \in \mathbb{R}^{n \times r}$ such that $\mathbf{C}\mathbf{C}^\top = \mathbf{B}^\top \mathbf{B} = \mathbf{I} \in \mathbb{R}^{r \times r}$ and*

$$\text{rank}(\mathbf{CAB}) = r. \quad (7)$$

*Proof.* Let $\mathbf{UDV}^\top$ be SVD of $\mathbf{A}$. Suppose $\mathbf{C} = \mathbf{U}_{1:r}^\top$ and $\mathbf{B} = \mathbf{V}_{1:r}$, where $\mathbf{U}_{1:r}$ and $\mathbf{V}_{1:r}$ are the matrices consisting of the singular vectors corresponding to the $r$ largest singular values. $\mathbf{C}$ and $\mathbf{B}$ satisfy $\text{rank}(\mathbf{CAB}) = r$, which concludes the proof. □

The constant $r$ can be represented in the matrix form with Lemma 1. Now, we show the characteristics of the presented solution by SVD in Lemma 1 with Lemma 2.

**Lemma 2.** *For any $\mathbf{u} = \{\mathbf{w}|\mathbf{w} \perp span\{\mathbf{u}_1, \cdots, \mathbf{u}_{k-1}\}\}$, $\mathbf{v} = \{\mathbf{w}|\mathbf{w} \perp span\{\mathbf{v}_1, \cdots, \mathbf{v}_{k-1}\}\}$ and $\mathbf{A} \in \mathbb{R}^{m \times n}$,*

$$\sigma_k = \max_{\mathbf{u},\mathbf{v}} \frac{|\mathbf{u}^\top \mathbf{A} \mathbf{v}|}{\|\mathbf{u}\| \|\mathbf{v}\|}. \quad (8)$$

Lemma 2 is the well-known *Variational Characterization of Singular Values* (or *Courant-Fischer Min-max principle for singular values*). By Lemma 2, we see that $\mathbf{C}$ and $\mathbf{B}$ satisfying Lemma 1 are also the unique solution of the following problem:

$$\max_{\mathbf{C},\mathbf{B}} \ \|\mathbf{CAB}\|_* \ \text{s.t.} \ \mathbf{CC}^\top = \mathbf{B}^\top \mathbf{B} = \mathbf{I}. \quad (9)$$

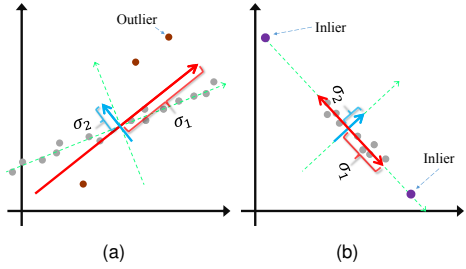

Fig. 1: Illustrations of the potential problems in minimizing nuclear norm (all singular values). The ground truth subspace (*green*) is a 1D line corrupted with sparse outliers and noise. In (a), the estimated subspace is biased to the estimated axis that has a smaller nuclear norm but a second singular value larger than the ground truth coordinate. In (b), some inliers located on the ground truth sub-space are regarded as outliers to achieve a smaller singular value.

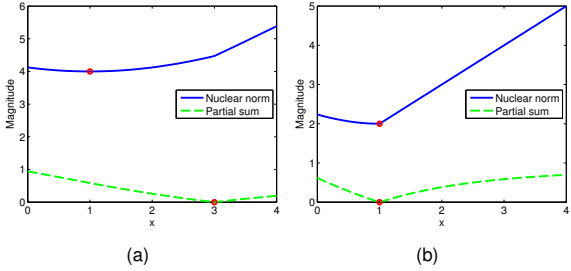

Fig. 2: A toy example comparison between the nuclear norm solution and the PSSV solution. Y-axis represents the magnitude of the nuclear norm and PSSV. The red dots represent the minimum points of the magnitude. The graphs show the nuclear norm and the PSSV of $2 \times 2$ matrices $\mathbf{A} = [1\,1; 3\,x]$ in (a) and $[1\,1; 1\,x]$ in (b) with $x$ varying from 0 to 4. In (a), the minimum point of nuclear norm is at $x = 1$ where the singular values of $\mathbf{A}$ are equal to $[3.4142, 0.5858]$ (i.e., rank-2). As for the PSSV, the minimum point is at $x = 3$ with singular values equal to $[4.4721, 0.0000]$ (i.e., rank-1). In this toy example, the nuclear norm favors a rank-2 solution over a rank-1 solution because the rank-2 solution provides the minimum nuclear norm. In contrast, the PSSV achieves the lowest rank (rank-1) solution. In (b), when the basis of the first row of $\mathbf{A}$ is partially supported by another sample (second row), the nuclear norm and the PSSV both achieve the rank-1 solution at minima.

Though the solution satisfying Lemma 1 is not unique, the solution of Eq. (9) can be a way to represent the target rank constraint. Therefore, we relax the target rank constant to the nuclear norm representation in Eq. (9) with $\mathbf{U}_{1:r}$ and $\mathbf{V}_{1:r}$ which satisfy both Lemma 1 and Eq. (9). In summary, we show the first term in Eq. (4) can be relaxed as PSSV by Lemmas 1 and 2, and the introduced projection operator $P_r(\cdot)$ (see Definition 1) favors preserving the information of the $r$ largest singular values. Intuitively, when $\text{rank}(\mathbf{A}) \geq r$, as $\sigma_{i>N}(\mathbf{A}) \to 0$ (namely minimized), $\text{rank}(\mathbf{A}) \to N$. Of course, if $\text{rank}(\mathbf{O}) < N$, i.e. if the inputs are degenerated, the rank of $\mathbf{A}$ cannot be increased to meet the target rank. This is a fundamental limitation. In common cases where input data contains noise or outliers, the inequality condition $\text{rank}(\mathbf{A}) \geq r$ is easily satisfied, because data corruptions increase the rank of input data.

### 3.1.2 Why the partial sum of singular values?

A major advantage of using the PSSV over the nuclear norm is that it does not minimize the variance of data distribution within the target rank. Minimizing the nuclear norm can favor a solution that has a lower nuclear norm, but the singular values in residual ranks (singular values above the target rank $N$. Let $N$=1 here) can be still large as illustrated in Fig. 1-(a) and Fig. 2-(a). This bias can degrade the accuracy of the estimated low-rank subspace. The bias phenomenon by a convex surrogate is common, and it could be corrected by non-convex relaxation [46]. An additional issue is that if the ground truth has a large variance but a sparse distribution within the ground truth subspace, some inliers can be regarded as outliers in order to reduce the singular values within the target rank, as illustrated in Fig. 1-(b) and at the minimum point of nuclear norm in Fig. 2-(a). These two problems are not an issue when there is a lot of observed data to support the correct estimation of $\mathbf{A}$. However, when observed data is limited, minimizing nuclear norm can easily cause bias since there is not a sufficient number of truth samples to support large variance of $\mathbf{A}$ within the target rank. In contrast, the PSSV does not assume small variance of $\mathbf{A}$, and it only minimizes variances in residual rank which corresponds to minimizing the noise variance of observed data. Note that the original rank operator, $\text{rank}(\mathbf{A})$, in Eq. (1) does not favor small variance solution.

### 3.2 Optimization by ADMM

Our partial sum objective function in Eq. (3) forms a constrained optimization problem. To solve this type of problems, Lin *et al.* [31] proposed an ADMM method (or called inexact augmented Lagrange multipliers, iALM). The augmented Lagrangian function of Eq. (3) is formulated by:

$$L_\mu(\mathbf{A}, \mathbf{E}, \mathbf{Z}) = \|\mathbf{A}\|_{p=N} + \lambda \|\mathbf{E}\|_1 \qquad (10)$$
$$+ \langle \mathbf{Z}, \mathbf{O} - \mathbf{A} - \mathbf{E} \rangle + \frac{\mu}{2} \|\mathbf{O} - \mathbf{A} - \mathbf{E}\|_F^2,$$

where $\mu$ is a positive scalar, $\mathbf{Z} \in \mathbb{R}^{m \times n}$ is an estimate of the Lagrange multiplier, $\|\cdot\|_F$ denotes the Frobenius norm, and $\langle \cdot, \cdot \rangle$ represents the inner product operator. Directly minimizing the Lagrangian function might be particularly challenging. According to a recent development of alternating direction method [31], Eq. (10) can be solved by minimizing each variable alternatively while fixing the other variables. The optimization problem can be divided into two sub-problems:

$$\mathbf{A}^* = \arg\min_{\mathbf{A}} L_{\mu_k}(\mathbf{A}, \mathbf{E}_k, \mathbf{Z}_k)$$
$$= \arg\min_{\mathbf{A}} \mu_k^{-1} \|\mathbf{A}\|_{p=N} + \frac{1}{2} \|\mathbf{A} - (\mathbf{O} - \mathbf{E}_k + \mu_k^{-1}\mathbf{Z}_k)\|_F^2, \qquad (11)$$

$$\mathbf{E}^* = \arg\min_{\mathbf{E}} L_{\mu_k}(\mathbf{A}_{k+1}, \mathbf{E}, \mathbf{Z}_k)$$
$$= \arg\min_{\mathbf{E}} \lambda \mu_k^{-1} \|\mathbf{E}\|_1 + \frac{1}{2} \|\mathbf{E} - (\mathbf{O} - \mathbf{A}_{k+1} + \mu_k^{-1}\mathbf{Z}_k)\|_F^2, \qquad (12)$$

where $k$ indicates the iteration index (see Alg. 1).

### 3.3 Solving $\mathbf{A}^*$

To minimize Eq. (11), we define the Partial Singular Value Thresholding (PSVT) operator $\mathbb{P}_{N,\tau}[\cdot]$. Before defining the PSVT, we first introduce the von Neumann's lemma (see details in de Sá *et al.* [14]).

**Lemma 3** (von Neumann [14]). *For any matrices $\mathbf{B}, \mathbf{Z} \in \mathbb{R}^{m \times n}$ and vectors of the singular values $\sigma(\cdot)$, the following*

*equality holds:*

$$\max\left\{\langle \mathbf{U}\mathbf{Z}\mathbf{V}^\top, \mathbf{B}\rangle \,|\, \mathbf{U} \in \mathcal{U}_m, \mathbf{V} \in \mathcal{U}_n\right\} = \langle \sigma(\mathbf{Z}), \sigma(\mathbf{B})\rangle, \quad (13)$$

*where $\mathcal{U}_n$ denotes the set of $n \times n$ unitary matrices, and $\langle \mathbf{A}, \mathbf{B}\rangle = \text{Tr}(\mathbf{A}^\mathrm{T}\mathbf{B})$, for any matrix $\mathbf{A} \in \mathbb{R}^{m \times n}$. Hence*

$$\langle \mathbf{A}, \mathbf{B}\rangle \leq \langle \sigma(\mathbf{A}), \sigma(\mathbf{B})\rangle. \quad (14)$$

*Moreover, equality holds in Eq. (14) iff there exists a simultaneous SVD $\mathbf{U}$ and $\mathbf{V}^\top$ of $\mathbf{A}$ and $\mathbf{B}$ in the following form:*

$$\mathbf{A} = \mathbf{U}\,\text{diag}\,(\sigma(\mathbf{A}))\,\mathbf{V}^\top \;\; and \;\; \mathbf{B} = \mathbf{U}\,\text{diag}\,(\sigma(\mathbf{B}))\,\mathbf{V}^\top. \quad (15)$$

The von Neumann's lemma shows that $\langle \mathbf{A}, \mathbf{B}\rangle$ is always bounded by the inner product of $\sigma(\mathbf{A})$ and $\sigma(\mathbf{B})$. Notice that the maximum value of $\langle \mathbf{A}, \mathbf{B}\rangle$ can be only achieved when $\mathbf{A}$ has the same singular vector matrices $\mathbf{U}$ and $\mathbf{V}$ as $\mathbf{B}$. This fact is useful to derive the PSVT.

**Theorem 1** (PSVT). *Let $\tau > 0$, $l = \min(m, n)$ and $\mathbf{X}, \mathbf{Y} \in \mathbb{R}^{m \times n}$ which can be decomposed by SVD. $\mathbf{Y}$ can be considered as the sum of two matrices, $\mathbf{Y} = \mathbf{Y}_1 + \mathbf{Y}_2 = \mathbf{U}_{Y1}\mathbf{D}_{Y1}\mathbf{V}_{Y1}^\top + \mathbf{U}_{Y2}\mathbf{D}_{Y2}\mathbf{V}_{Y2}^\top$, where $\mathbf{U}_{Y1}, \mathbf{V}_{Y1}$ are the singular vector matrices corresponding to the N largest singular values by SVD, and $\mathbf{U}_{Y2}, \mathbf{V}_{Y2}$ from the (N+1)-th to the last singular values. Define a minimization problem for the PSSV as*

$$\arg\min_{\mathbf{X}} \; \frac{1}{2}\|\mathbf{X} - \mathbf{Y}\|_F^2 + \tau \|\mathbf{X}\|_{p=N}. \quad (16)$$

*Then, the optimal solution of Eq. (16) can be expressed by the PSVT operator defined as:*

$$\begin{aligned}\mathbb{P}_{N,\tau}[\mathbf{Y}] &= \mathbf{U}_Y(\mathbf{D}_{Y1} + \mathcal{S}_\tau[\mathbf{D}_{Y2}])\mathbf{V}_Y^\top \\ &= \mathbf{Y}_1 + \mathbf{U}_{Y2}\mathcal{S}_\tau[\mathbf{D}_{Y2}]\mathbf{V}_{Y2}^\top,\end{aligned} \quad (17)$$

*where*

$$\begin{aligned}\mathbf{D}_{Y1} &= \text{diag}(\sigma_1, \cdots, \sigma_N, 0, \cdots, 0), \\ \mathbf{D}_{Y2} &= \text{diag}(0, \cdots, 0, \sigma_{N+1}, \cdots, \sigma_l),\end{aligned}$$

*and $\mathcal{S}_\tau[x] = \text{sign}(x) \cdot \max(|x| - \tau, 0)$ is the soft-thresholding operator [17], [22].*

*Proof.* Let's consider $\mathbf{X} = \mathbf{U}_X \mathbf{D}_X \mathbf{V}_X^\top = \sum_{i=1}^{l} \sigma_i(\mathbf{X}) u_i v_i^\top$ where $\mathbf{U}_X = [u_1, \cdots, u_m] \in \mathcal{U}_m$, $\mathbf{V}_X = [v_1, \cdots, v_n] \in \mathcal{U}_n$ and $\mathbf{D}_X = \text{diag}(\sigma(\mathbf{X}))$, where the singular values $\sigma(\cdot) = [\sigma_1(\cdot), \cdots, \sigma_l(\cdot)] \geq 0$ are sorted in a non-increasing order. Also we define the function $J(\mathbf{X})$ as the objective function of Eq. (16). The first term of Eq. (16) can be derived as follows:

$$\begin{aligned}\frac{1}{2}\|\mathbf{X} - \mathbf{Y}\|_F^2 &= \frac{1}{2}\left(\|\mathbf{Y}\|_F^2 - 2\langle \mathbf{X}, \mathbf{Y}\rangle + \|\mathbf{X}\|_F^2\right) \\ &= \frac{1}{2}\left(\|\mathbf{Y}\|_F^2 - 2\sum_{i=1}^l \sigma_i(\mathbf{X}) u_i^\top \mathbf{Y} v_i + \sum_{i=1}^l \sigma_i(\mathbf{X})^2\right) \\ &= \frac{1}{2}\|\mathbf{Y}\|_F^2 + \frac{1}{2}\sum_{i=1}^l \left(-2\sigma_i(\mathbf{X}) u_i^\top \mathbf{Y} v_i + \sigma_i(\mathbf{X})^2\right).\end{aligned} \quad (18)$$

In the minimization of Eq. (18) with respect to $\mathbf{X}$, $\|\mathbf{Y}\|_F^2$ is regarded as a constant and thus can be ignored. For a more detailed representation, we change the parameterization of $\mathbf{X}$ to $(\mathbf{U}_X, \mathbf{V}_X, \mathbf{D}_X)$ and minimize the function:

$$\begin{aligned}J(\mathbf{U}_X, \mathbf{V}_X, \mathbf{D}_X) &= \\ \frac{1}{2}\sum_{i=1}^l &\left(-2\sigma_i(\mathbf{X}) u_i^\top \mathbf{Y} v_i + \sigma_i(\mathbf{X})^2\right) + \tau \sum_{i=N+1}^l \sigma_i(\mathbf{X}).\end{aligned} \quad (19)$$

From von Neumann's lemma, the upper bound of $u_i^\top \mathbf{Y} v_i$ is given as $\sigma_i(\mathbf{Y}) = \max\{u_i^\top \mathbf{Y} v_i\}$ for all $i$ when $\mathbf{U}_X = \mathbf{U}_Y$ and $\mathbf{V}_X = \mathbf{V}_Y$. The lower envelope of $J(\mathbf{U}_X, \mathbf{V}_X, \mathbf{D}_X)$ is obtained at $\mathbf{U}_X = \mathbf{U}_Y$ and $\mathbf{V}_X = \mathbf{V}_Y$. Then Eq. (19) becomes a function depending only on $\mathbf{D}_X$ as follows:

$$\begin{aligned}&J(\mathbf{U}_Y, \mathbf{V}_Y, \mathbf{D}_X) \\ &= \frac{1}{2}\sum_{i=1}^l \left(-2\sigma_i(\mathbf{X})\sigma_i(\mathbf{Y}) + \sigma_i(\mathbf{X})^2\right) + \tau \sum_{i=N+1}^l \sigma_i(\mathbf{X}) \\ &= \frac{1}{2}\Bigg(\sum_{i=1}^N \left(-2\sigma_i(\mathbf{X})\sigma_i(\mathbf{Y}) + \sigma_i(\mathbf{X})^2\right) \\ &\quad + \sum_{i=N+1}^l \left(-2\sigma_i(\mathbf{X})\sigma_i(\mathbf{Y}) + \sigma_i(\mathbf{X})^2 + 2\tau\sigma_i(\mathbf{X})\right)\Bigg).\end{aligned} \quad (20)$$

Since Eq. (20) consists of simple quadratic equations for each $\sigma_i(\mathbf{X})$ independently, it is trivial to show that the minimum of Eq. (20) is obtained at $\hat{\mathbf{D}}_X = \text{diag}(\hat{\sigma}(\mathbf{X}))$ by derivative in a feasible domain as the first-order optimality condition, where $\hat{\sigma}_i(\mathbf{X})$ is defined as

$$\hat{\sigma}_i(\mathbf{X}) = \begin{cases} \sigma_i(\mathbf{Y}), & \text{if } i < N+1, \\ \max(\sigma_i(\mathbf{Y}) - \tau, 0), & \text{otherwise.} \end{cases} \quad (21)$$

Hence, the solution of Eq. (16) is $\mathbf{X}^* = \mathbf{U}_Y \hat{\mathbf{D}}_X \mathbf{V}_\mathbf{Y}^T$. This result exactly corresponds to the PSVT operator where a feasible solution $\mathbf{X}^* = \mathbf{U}_Y(\mathbf{D}_{Y1} + \mathcal{S}_\tau[\mathbf{D}_{Y2}])\mathbf{V}_Y^\top$ exists. □

Our proposed PSVT can be regarded as a special case of solving the weighted nuclear norm based objective function of Chen *et al.* [11] and Gaïffas *et al.* [20]. But we would like to notice that our method suggests how the weighted parameter (defined in their literatures) can be determined to encourage the rank constraint. Also, notice that our proposed PSVT operator provides a closed-form solution for systems of the same form as Eq. (16) (e.g. Eq. (11)). While Eq. (11) is a non-convex function, the PSVT provides a global optimal solution for the sub-problem of $\mathbf{A}$ (see the proof of Theorem 1).

As an analysis of PSVT, when $\tau = \infty$, the optimal solution of Eq. (16) is a low-dimensional projection of $\mathbf{Y}$ known as singular value projection [27] which enforces the target rank constraint through projection. When $\sigma_i < \tau$ for $1 \leq i \leq N$, conventional SVT [8] projects these $\sigma_i$ to zero resulting in a more deficient rank of $\mathbf{A}$ than the target rank while our PSVT does not lead to rank deficient matrices. Hence, PSVT implicitly encourages the resulting matrix $\mathbf{A}$ to meet the target rank even when all the $\sigma_i$ are small, which occasionally happens when the number of observed samples is limited.

## 3.4 Solving $\mathbf{E}^*$

As suggested by Hale *et al.* [22], the solution to the sub–problem in Eq. (12) can be obtained as:

$$\mathcal{S}_\tau[\mathbf{Y}] = \arg\min_{\mathbf{X}} \frac{1}{2}\|\mathbf{X} - \mathbf{Y}\|_F^2 + \tau\|\mathbf{X}\|_1, \quad (22)$$

where $\mathcal{S}_\tau[x] = \text{sign}(x)\max(|x|-\tau, 0)$ is the soft-thresholding operator [17], [22], and $x \in \mathbb{R}$. This operator can be extended to vectors and matrices by applying it element-wise. The soft-thresholding (shrinkage) method is shown to be very effective in minimizing $l^1$-norm and the proximity term, and guarantees that the solution is the global minimum for the equations of the same form as Eq. (22) (e.g. Eq. (12)) [17], [22].

## 3.5 Updating $\mathbf{A}^*$ and $\mathbf{E}^*$

At each iteration $k$, $\mathbf{A}_k$ and $\mathbf{E}_k$ can be updated with the operators $\mathcal{S}_\tau[\cdot]$ and $\mathbb{P}_{N,\tau}[\cdot]$ as:

$$\begin{aligned}\mathbf{A}_{k+1} &= \mathbb{P}_{N,\mu_k^{-1}}[\mathbf{O} - \mathbf{E}_k + \mu_k^{-1}\mathbf{Z}_k], \\ \mathbf{E}_{k+1} &= \mathcal{S}_{\lambda\mu_k^{-1}}[\mathbf{O} - \mathbf{A}_{k+1} + \mu_k^{-1}\mathbf{Z}_k].\end{aligned} \quad (23)$$

The iterations are terminated when the equality constraint is satisfied (in all the experiments, $\frac{\|\mathbf{O}-\mathbf{A}-\mathbf{E}\|_F}{\|\mathbf{O}\|_F} < 1e^{-7}$). Experiments showed that updating $\mathbf{A}_k$ and $\mathbf{E}_k$ for only one iteration in the inner loop is sufficient to produce a satisfying accurate solution of Eq. (3). This method is called the inexact ALM [31] and is designed for computational efficiency.

We summarize the overall algorithm in Alg. 1 (For more details, refer to the report of Lin *et al.* [31]).

---

**Algorithm 1** ADMM for the PSSV based RPCA

---

**Input :** $\mathbf{O} \in \mathbb{R}^{m\times n}, \lambda > 0$, the constraint rank $N$.
Initialize $\mathbf{A}_0 = \mathbf{E}_0 = \mathbf{0}$, $\mathbf{Z}$ as suggested in [31], $\mu_0 > 0$, $\rho > 1$ and $k = 0$.
// Outer loop
**while** not converged **do**
    // Inner loop
    **while** not converged **do**
        $\mathbf{A}_{k+1} = \mathbb{P}_{N,\mu_k^{-1}}[\mathbf{O} - \mathbf{E}_k + \mu_k^{-1}\mathbf{Z}_k]$.
        $\mathbf{E}_{k+1} = \mathcal{S}_{\lambda\mu_k^{-1}}[\mathbf{O} - \mathbf{A}_{k+1} + \mu_k^{-1}\mathbf{Z}_k]$.
    **end while**
    $\mathbf{Z}_{k+1} = \mathbf{Z}_k + \mu_k(\mathbf{O} - \mathbf{A}_{k+1} - \mathbf{E}_{k+1})$.
    $\mu_{k+1} = \rho\mu_k$.
    $k = k+1$.
**end while**
**Output :** $(\mathbf{A}_k, \mathbf{E}_k)$.

---

## 3.6 Convergence Analysis

To the best of our knowledge, the general convergence property of ADMM which alternates between non-convex (solving $\mathbf{A}^*$) and convex (solving $\mathbf{E}^*$) functions has not been answered yet. The ADMM for non-convex problems can be considered as a local optimization method, which aims to converge to a point with *better objective value* [4].

In our problem, each sub-problem has a closed-form solution and the objective value is always decreasing with respect to the primal variables optimized in each sub-problem iteration[2]. Our empirical convergence tests showed that our ADMM based algorithm has a strong convergence behavior (see Sec. 4.1). Although the global optimal solution is not guaranteed, all of our experiments showed that our algorithm converges to a solution which is very close to the nuclear norm solution, when the number of observations is more than sufficient. It also converges to a better solution than the nuclear norm solution when the number of observations is limited, even with all zero initializations.

---

2. It does not mean a monotonic decrease of the Lagrangian function, which is not necessarily monotone due to the dual update.

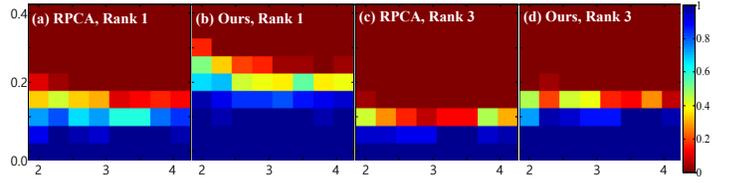

Fig. 4: Success ratio for synthetic data with a varying number of rows (dimension) $m$ (a-d). Comparison between RPCA (nuclear norm) and ours (PSSV) for the rank-1 case (a,b), and for the rank-3 case (c,d). Y–axis represents the corruption ratio $r \in [0, 0.4]$. X–axis represents the log scale row size $\log_{10} m \in [\log_{10} 100, \log_{10} 12800]$ in (a-d). The color magnitude represents the success ratio $[0,1]$.

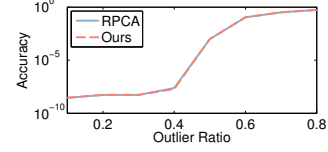

Fig. 5: NRMSE comparison on a sufficient sample condition with a rank-3 matrix $\mathbf{O} \in \mathbb{R}^{10000\times 3000}$. Under the sufficient sample case, the nuclear norm and PSSV solutions are very similar.

Besides the empirical behavior, we provide the convergence property for Alg. 1 in Proposition 1. It shows that any accumulation point (limit point) generated along the iterations satisfies the first-order necessary optimal condition, a KKT (Karush-Kuhn-Tucker) point.

**Proposition 1** (Convergence). *Let $S_k = (\mathbf{A}_k, \mathbf{E}_k, \mathbf{Y}_k, \hat{\mathbf{Y}}_k)$, where $\hat{\mathbf{Y}}_{k+1} = \mathbf{Y}_k + \mu_k(\mathbf{O} - \mathbf{A}_{k+1} - \mathbf{E}_k)$ and $\{S_k\}_{k=1}^\infty$ is a set of intermediate solutions of Alg. 1. Suppose that $\{\mathbf{Y}_k\}_{k=1}^\infty$ and $\{\hat{\mathbf{Y}}_k\}_{k=1}^\infty$ are bounded, $\lim_{k\to\infty}(\mathbf{Y}_{k+1} - \mathbf{Y}_k) = 0$, and $\mu_k$ is non-decreasing, then any accumulation point of $\{S_k\}_{k=1}^\infty$ satisfies the following KKT conditions: (C1) $\mathbf{Y}^* \in \partial^C \|\mathbf{A}^*\|_p$, (C2) $\mathbf{Y}^* \in \partial\|\lambda\mathbf{E}^*\|_1$, (C3) $\mathbf{O} - \mathbf{A}^* - \mathbf{E}^* = \mathbf{0}$, (C4) $\partial^C \|\mathbf{A}^*\|_p \cap \partial\|\lambda\mathbf{E}^*\|_1 \neq \emptyset$. In particular, whenever $\{S_k\}_{k=1}^\infty$ converges, it converges to a KKT point of Eq. (3).*

The proofs can be found in the supplementary material. The conditions for non-decreasing $\mu_k$ and the boundness of the sequence are already satisfied by Alg. 1 (see Lemma 1 in Lin *et al.* [31]). Proposition 1 is established for a single iteration algorithm in the inner loop, i.e. iALM. When the inner loop iterates until convergence (exactly solving the inner loop), it may lead a simpler proof than the above result. We remain further theoretical analyses of convergence as future work.

## 4 EXPERIMENT RESULTS

We compare the performance of the proposed method against RPCA (nuclear norm) [9] with synthetic data sets and real world application examples. In all the experiments, we use the default parameters recommended by Candès *et al.* [9] for both their approach and ours, i.e. $\lambda = 1/\sqrt{max(m,n)}$ and $\rho = 1.5$, except if explicitly stated

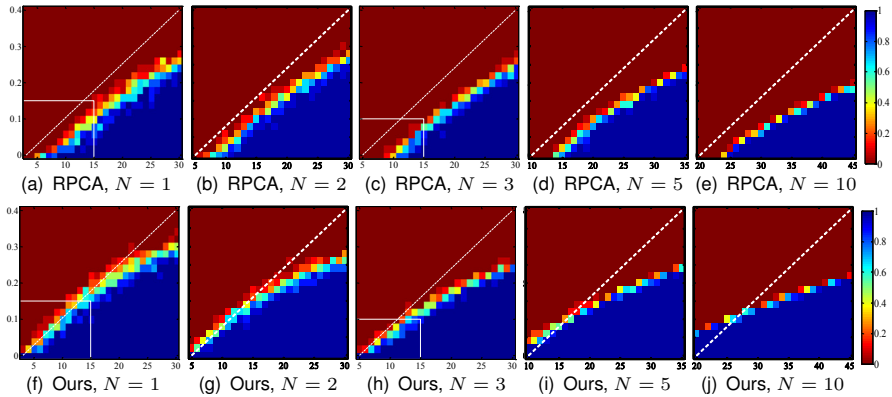

Fig. 3: Success ratio for synthetic data with varying the number of columns (observations) $n$. Comparison between RPCA (nuclear norm) and ours (PSSV) for rank-1,2,3,5,10 cases. X–axis represents the column size, and Y–axis represents the corruption ratio $r \in [0, 0.4]$. The color magnitude represents the success ratio $[0,1]$. The white dotted lines are provided as a guide for easier comparison.

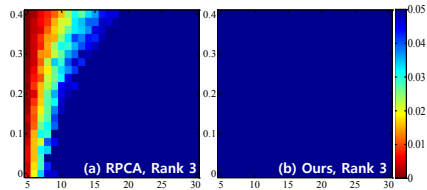

Fig. 6: Comparison for the rank deficiency of the estimated low-rank matrix $\hat{\mathbf{A}}$ for the rank-3 case, obtained by RPCA (a) and our method (b). The red regions indicate rank deficiency, i.e. the rank of the recovered matrix is lower than the constraint rank. X–axis represents the column size, and Y–axis represents the corruption ratio $r \in [0, 0.4]$. The color magnitude represents the success ratio $[0,1]$.

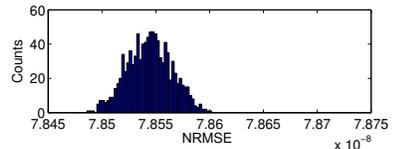

Fig. 7: Distribution of residual errors with 1000 different random initializations.

otherwise. The code of the proposed method is available on our project website [3].

### 4.1 Synthetic Dataset

We compare our method (PSSV) with RPCA (nuclear norm) on synthetic data by evaluating the success ratio and convergence behaviors. To synthesize a ground-truth low-rank matrix $\mathbf{A}_{GT} \in \mathbb{R}^{m \times n}$ of rank-$N$, we perform a linear combination of $N$ arbitrary orthogonal basis vector. The weight vector used to span each column vector of $\mathbf{A}_{GT}$ is randomly sampled from the uniform distribution $U[0, 1]$. To generate sparse outliers, we select $m \times n \times r$ entries from $\mathbf{A}_{GT}$, where $r$ denotes the corruption ratio. Larger $r$ means more outlier entries. The selected entries are corrupted by random noise from $U[0, 1]$. We run each of the tests over 50 trials and report the overall average errors of the trials. We refer to $\frac{\|\mathbf{A}_{GT} - \hat{\mathbf{A}}\|_F}{\|\mathbf{A}_{GT}\|_F}$ as the normalized root mean squared error (NRMSE).

#### 4.1.1 Comparison of Success Ratio

We verify the robustness of RPCA (nuclear norm) and the proposed method (PSSV) with respect to the number of observations, data dimension and corruption ratio. We examine the performance by counting the number of successes. If the recovered $\hat{\mathbf{A}}$ has a NRMSE smaller than 0.01, we consider the estimation of $\mathbf{A}$ and $\mathbf{E}$ is successful. We compare the success ratio with varying column size $n$ (i.e. the number of observations), and row size $m$ (i.e. data dimension). The magnitude in Fig. 3 indicates the success

3. http://thoh.kaist.ac.kr

percentage. A larger blue area indicates a more robust performance of the algorithm.

We also perform experiments where we fix $m = 10,000$ and vary $n$ and $r$. The comparison between RPCA and our method with rank-1,2,3,5 and -10 constraints is shown in Fig. 3. As $n$ decreases (i.e. the number of observations decreases), the success ratio of RPCA decreases more rapidly than our method. When more observations are available (over $n = 25$ in Fig. 3 and Fig. 5), both methods show a similar behavior.

Fig. 4 shows the success ratio of RPCA and ours for the varying row $m$ cases. We fix $n = 16$, and vary $m$ and $r$. Our method can successfully recover $\mathbf{A}$ and $\mathbf{E}$ with input data contaminated up to $15\%$ of severe corruption for the rank-1 case in Fig. 4-(b), and leads to more robust results than RPCA despite $5\%$ higher corruption for the rank-3 case in Fig. 4-(d).

#### 4.1.2 Rank Deficiency

We verify whether the recovered $\hat{\mathbf{A}}$ obtained by RPCA and our method is rank deficient. Our objective function minimizes the rank of $\mathbf{A}$ up to the target rank. Thus, the result rank of $\hat{\mathbf{A}}$ should not be lower than the target rank. In practice, rank deficiency is crucial for quality of the final solution in some applications (e.g. photometric stereo). We measure the ratio $\sigma_N(\hat{\mathbf{A}})/\sigma_1(\hat{\mathbf{A}})$ (similar to the inverse value of the condition number) for the rank-$N$ constraint case. We only test for rank-$N = 3$ as a typical example of photometric stereo. If the ratio is lower than 0.01, we consider that the recovered matrix has a rank lower than $N$. In Fig. 6, the red regions mean that the rank of the recovered matrix is lower than the target rank. The experiments empirically validate that the rank obtained by our method is bounded for almost all of the regions, while RPCA has regions whose rank is lower than the target rank. This happens when observations do not support its true subspaces well.

| Method | Objective function | Constraint |
|---|---|---|
| Eriksson *et al.* [18] | $\min_{\mathbf{U},\mathbf{V}} \|\mathbf{O}-\mathbf{UV}\|_1$ | – |
| Zheng *et al.* [48] | $\min_{\mathbf{U},\mathbf{V}} \|\mathbf{O}-\mathbf{UV}\|_1 + \lambda\|\mathbf{V}\|_*$ | $\mathbf{U}^\top\mathbf{U}=\mathbf{I}$ |
| LMaFit [40] | $\min_{\mathbf{A},\mathbf{U},\mathbf{V}} \|\mathbf{O}-\mathbf{A}\|_1$ | $\mathbf{A}=\mathbf{UV}$ |
| SVP [27] based RPCA | $\min_{\mathbf{A},\mathbf{E}} \|\mathbf{E}\|_1$ | $\mathbf{O}=\mathbf{A}+\mathbf{E}$, $\mathrm{rank}(\mathbf{A})=N$ |
| RPCA [9] | $\min_{\mathbf{A},\mathbf{E}} \|\mathbf{A}\|_* + \lambda\|\mathbf{E}\|_1$ | $\mathbf{O}=\mathbf{A}+\mathbf{E}$ |
| WNNM [11] based RPCA | $\min_{\mathbf{A},\mathbf{E}} \|\mathbf{A}\|_{\mathbf{w},*} + \lambda\|\mathbf{E}\|_1$ | $\mathbf{O}=\mathbf{A}+\mathbf{E}$ |
| Our method | $\min_{\mathbf{A},\mathbf{E}} \|\mathbf{A}\|_{p=N} + \lambda\|\mathbf{E}\|_1$ | $\mathbf{O}=\mathbf{A}+\mathbf{E}$ |

TABLE 1: Summary of the compared methods. $\mathbf{O},\mathbf{A},\mathbf{E} \in \mathbb{R}^{m\times n}, \mathbf{U}\in \mathbb{R}^{m\times N}, \mathbf{V}\in \mathbb{R}^{N\times n}$, $\|\cdot\|_{\mathbf{w},*}$ is the weighted nuclear norm and the weight coefficients in $\mathbf{w}$ is determined adaptively as suggested by Chen *et al.* [11].

### 4.1.3 Sensitivity to Initialization

Since the proposed objective function is non-convex, the converged solution may be different according to the initialization. To study the sensitivity of the optimization against the initialization, we conducted 1000 experiments with random initialization on a rank-3 matrix $\mathbf{O}\in\mathbb{R}^{10000\times 50}$ with 5% outliers. The distribution of NRMSE is shown in Fig. 7. While the convergence of non-convex problem to an optimum is hard to be guaranteed, most solutions are concentrically distributed in regions near the ground-truth solution with small errors.

### 4.1.4 Comparisons with other low-rank approximations

We provide additional comparisons with the singular value projection (SVP) [27] based and the weighted nuclear norm (WNNM) [11] based methods, and low-rank matrix approximation approaches by MF. The formulations are summarized in Table 1. The SVP and WNNM are reformulated based on RPCA framework for fair comparison. MF methods enforce the target rank $N$ constraint of data matrix ($\mathbf{O}=\mathbf{UV}$) by factorizing it into a product of rank-$N$ basis ($\mathbf{U}$) and coefficient ($\mathbf{V}$) as hard constraint. Among the existing MF based methods, we compare with the state-of-the-art methods of LMaFit [40], Zheng *et al.* [48] and Eriksson *et al.* [18], with the default recommended parameters.

Since the method of Eriksson *et al.* can only handle small size examples, we perform separate experiments for small and large scales. We synthetically generate data matrices $\mathbb{R}^{30\times 7}$ with rank-2 for small scale or $\mathbb{R}^{5000\times 20}$ with rank-3 for large scale, and varying corruption ratio in $[0.05, 0.20]$. NRMSE is displayed in Fig. 8-(a,b).

Compared to our method, their approach also minimizes the nuclear norm in addition to the hard target rank constraint. As discussed previously, since minimizing the nuclear norm also implicitly minimizes the variance of the estimated low-rank matrix, their estimated low-rank matrix could be biased by this assumption. On the other hand, since our PSSV objective function does not have this assumption, and since the target rank is penalized softly, our method converges to more accurate solutions compared to the solutions of LMaFit, Zheng *et al.* and Eriksson *et al.*

We have also conducted experiments for the under-sampled cases on subspace: e.g. a ground truth data is spanned with 3 basis axis (true and target rank are 3), but the distribution along the third basis axis has a very small variance (small singular value). Thus, although the underlying matrix is a rank-3 matrix, it is very close to a rank-2 matrix. This situation often happens when the last basis is less supported by a few true samples. This is also called the unbalanced singular values case (e.g. $\sigma(\mathbf{A})=[100,10,1e^{-1}]$). Our results shown in Fig. 8-(c,d) have smaller errors than results from LMaFit, Zheng *et al.* and Eriksson *et al.*, even for the under-sampled cases.

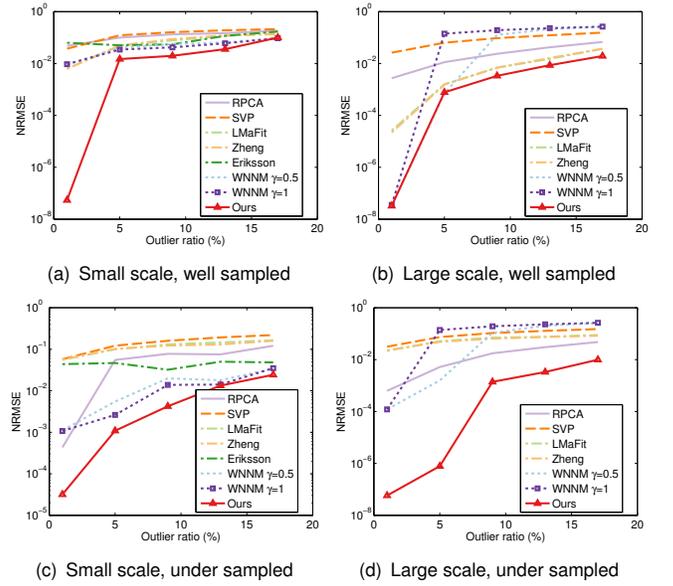

(a) Small scale, well sampled  (b) Large scale, well sampled

(c) Small scale, under sampled  (d) Large scale, under sampled

Fig. 8: Accuracy comparisons with varying outlier ratio and deficient number of samples for SVP [27] based and WNNM [11] based methods, LMaFit [40], Zheng *et al.* [48], Eriksson *et al.* [18], and our method. The experiments consist of small scale problems ($\mathbf{O}\in\mathbb{R}^{30\times 7}$ with rank-2) in (a,c) and large scale problems ($\mathbf{O}\in\mathbb{R}^{5000\times 20}$ with rank-3) in (b,d). The cases with well-sampled and under-sampled data on subspaces are shown at the top and bottom rows respectively. X-axis represents the percentage of outlier, and Y-axis represents the average error. LMaFit, Zheng *et al.* and Eriksson *et al.* are MF methods. MF methods also result in low accuracy under the case of deficient number of samples. Comparing (b) and (d), MF methods are prone to the data under-sampled on subspaces, because bilinear model enforcedly constrains the target rank and excessively attempts to match it.

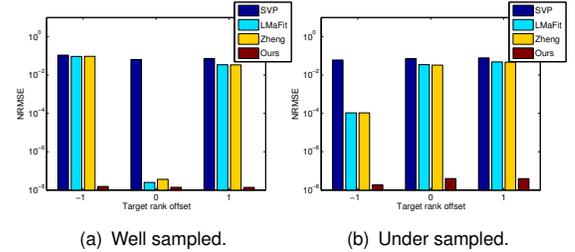

(a) Well sampled.  (b) Under sampled.

Fig. 9: Effects when the target rank is incorrectly set. We set the input target rank $N=N_{\text{true}}+t_{\text{offset}}$, where the truth rank $N_{\text{true}}=3$. The lower value the better.

### 4.1.5 Incorrect Setting of Target Rank

Our method takes advantage of the target rank from the problem definition. When the target rank is set incorrectly, the question of the behavior of our method naturally arises. For the sake of completeness, we have experimented with incorrect target rank setting in Fig. 9.

We considered the situation where the rank is known, but ambiguous within some bound (e.g., the truth rank is 3, but ambiguous within rank-{2,3,4}). The data construction is similar to the experiments conducted in Sec. 4.1.4, i.e. well-sample and under-sampled data cases. The rank-3 matrices $\mathbf{O}\in\mathbb{R}^{3000\times 100}$ are used for experiment. Fig. 9 shows that MF based methods are prone to incorrect target rank setting. Interestingly, for the data under-sampled on subspaces, MF based methods with a target rank lower than the true rank show better performance than for the well-sampled data case. This is because the bilinear model enforcedly constrains the target rank within the bilinear matrix structure. Therefore, when the $3^{\text{rd}}$ basis is weakly supported by samples, fitting with a rank-2 bilinear model only for the $1^{\text{st}}$ and $2^{\text{nd}}$ basis provides better precision than using a rank-3 bilinear model. This result is consistent with Fig. 8-(b,d).

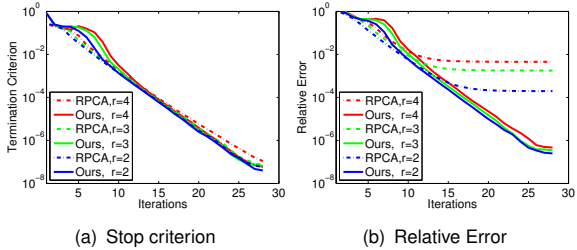

(a) Stop criterion     (b) Relative Error

Fig. 10: Convergence behavior of RPCA [31] and our method for the rank 2,3 and 4.

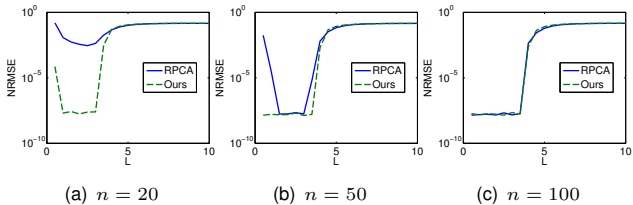

(a) $n = 20$     (b) $n = 50$     (c) $n = 100$

Fig. 11: Evolution of the NRMSE according to varying $\lambda = L/\sqrt{\min(m,n)}$. For experiments, rank-3 matrices $\mathbf{O} \in \mathbb{R}^{5000 \times n}$ are generated with $5\%$ outliers.

#### 4.1.6 Convergence Behavior

To examine the convergence behavior of both RPCA [31] and our method, we plot the evolution of the relative errors $\frac{\|\mathbf{A}_{GT}-\hat{\mathbf{A}}\|_F}{\|\mathbf{A}_{GT}\|_F} + \frac{\|\mathbf{E}_{GT}-\hat{\mathbf{E}}\|_F}{\|\mathbf{E}_{GT}\|_F}$ and termination criteria $\frac{\|\mathbf{O}-\mathbf{A}-\mathbf{E}\|_F}{\|\mathbf{O}\|_F}$ over the iterations in Fig. 10-(a) and (b), respectively. We randomly generate $5000 \times 40$ matrices for the rank-$2,3,4$ cases, and the average value over the trials is computed.

We use the *MATLAB* implementation of RPCA provided by Wright *et al.* [44]. We run our method until convergence, and we observe that it is terminated at similar moments with RPCA as shown in Fig. 10-(a). Also, our method takes the same amount of time as inexact ALM based RPCA [31]. Fig. 10-(b) also shows that our method provides higher accuracy than RPCA as well as a gradual convergence under the same termination criterion.

#### 4.1.7 Lambda ($\lambda$) parameter

We conduct all the experiments in this paper with the same $\lambda$ parameter recommended by Candès *et al.* [9]. For completeness, we show in this section how the choice of $\lambda$ can affect the solution of both RPCA and ours. Note that tuning the optimal $\lambda$ to balance the nuclear norm and sparsity is not possible unless the ground truth solution is known as discussed by Chandrasekaran *et al.* [10]. Thus, the results provided here are only for reference. Fig. 11 shows normalized MSE when $\lambda$ varies, where $\lambda = L/\sqrt{max(m,n)}$. The results show that our method consistently produces less errors than RPCA under different settings of $\lambda$.

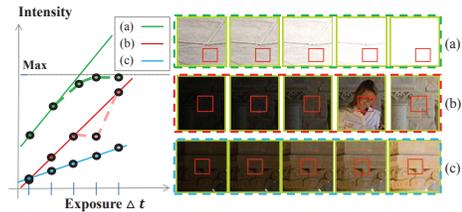

Fig. 12: Illustration of the observed intensity values for (a) saturation region, (b) moving object, and (c) consistent cases. Solid lines denote ideal relationship between intensity and exposure, and dots and dotted lines denote the observed intensities.

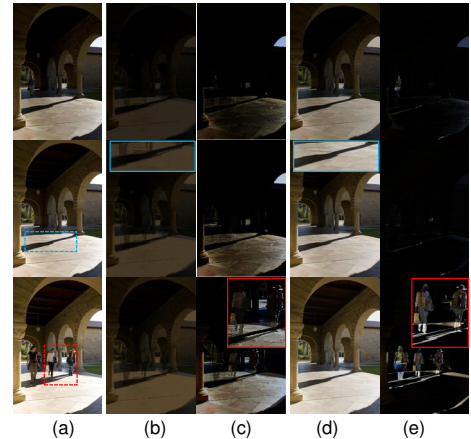

(a)    (b)    (c)    (d)    (e)

Fig. 13: Comparison of the low-rank matrix and sparse error results between RPCA and ours on the *Arch* dataset [21]. (a) Three out of the five input multi–exposure image samples. Low-rank (b,d) and sparse error (c,e) results, respectively obtained by RPCA (b,c) and the proposed approach (d,e).

### 4.2 Real-world Applications

#### 4.2.1 High Dynamic Range (HDR) Imaging

We apply the proposed method for modeling a background scene and a ghost-free HDR composition. The input is a set of low dynamic range (LDR) images and the goal is to composite an HDR image using RPCA to reject outliers, such as moving objects and saturations, in the LDR images. We assume that the differently exposed images $I_i$ are aligned and the camera response function (CRF) is calibrated (or linear). Then, the captured images can be represented as $I_i = \kappa R \Delta t_i$, where $R$ denotes the sensor irradiance, $\Delta t_i$ is the exposure time of the $i$-th image, and $\kappa$ is a positive scalar. We construct the observed intensity matrix $\mathbf{O} \in \mathbb{R}^{m \times n} = [\text{vec}(I_1)| \cdots |\text{vec}(I_n)]$ by stacking the vectorized input images, where $m$ and $n$ are the number of pixels and images respectively. Since the intensities of the input images are linearly dependent, the ideal solution of this problem is rank-$1$. However, in practice, $\text{rank}(\mathbf{O})$ is higher than $1$ due to moving objects, saturation or other artifacts (illustrated in Fig. 12). We apply RPCA (nuclear norm) and our method (PSSV) to each color channel independently, in order to separate artifacts and background scene.

The *Arch* and *Sculpture Garden* datasets from Gallo *et al.* [21] are used for evaluation. The estimated backgrounds as low-rank matrix and the sparse outliers from RPCA and our method are shown in Fig. 13. The example in Fig. 13-(a) consists of only 5 input images which is very limited. Ideally, the decomposed low-rank matrix $\mathbf{A} = [\text{vec}(A_1)| \cdots |\text{vec}(A_n)]$ consists of relative intensities of the

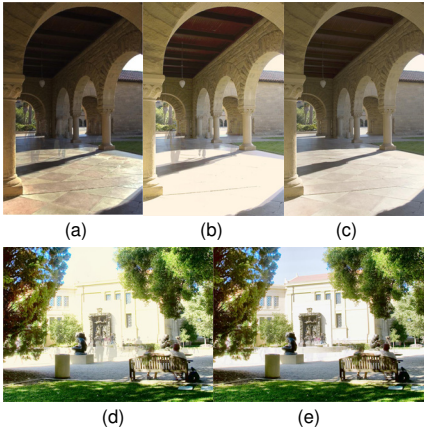

Fig. 14: HDR composition results for the Arch (top) and Sculpture Garden (bottom) datasets [21]. (a) Debevec *et al*. [15]. (b,d) RPCA [31]. (c,e) the proposed method.

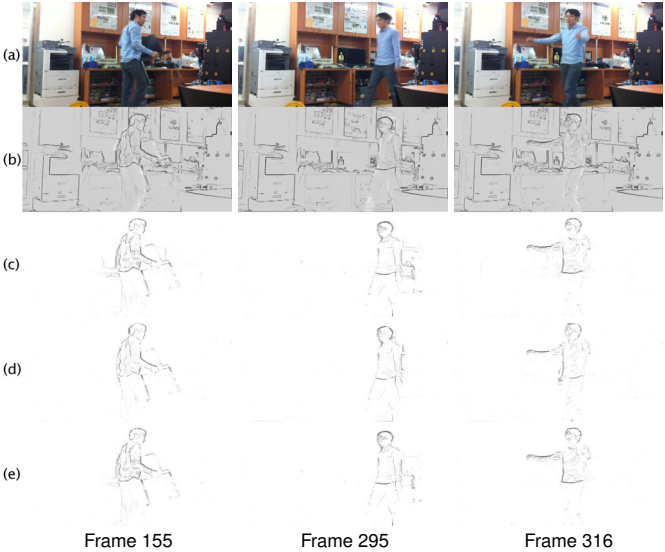

Frame 155     Frame 295     Frame 316

Fig. 15: Motion detection comparison between RPCA and our method on a varying illumination dataset [34]. (a) Representative sampled inputs. (b-c) Results of RPCA, with $n = 5$ and $100$ respectively. (d-e) Our results, with $n = 5$ and $100$ respectively.

background scene from which moving objects or saturation artifacts should be removed (see Fig. 13-(b,d)). RPCA returns a low-rank matrix whose magnitude differs drastically from the input image, as shown in Fig. 13-(b). Moreover RPCA yields a dense non-zero entries in $\mathbf{E}$, instead of being sparse, as shown in Fig. 13-(c). This situation is similar to the example in Fig. 2 where the minimum nuclear norm favors a solution with smaller variance of magnitudes. In contrast, our proposed method shows a correctly modeled background scene and successfully detects outlier regions, as shown in Fig. 13-(d,e). For displaying the sparse components in Fig. 13-(c,e), each color component (R,G,B) is set with $(|\mathbf{E}_R|, |\mathbf{E}_G|, |\mathbf{E}_B|)$, where $\mathbf{E}_{\{R,G,B\}}$ denotes sparse error matrix for each channel.

After we estimate the low-rank matrix, we composite the HDR images using the standard method of Debevec *et al*. [15]. The final HDR results are shown in Fig. 14. Because the background modelling by RPCA is inaccurate, ghosting appears in their HDR results. In contrast, our results are ghost-free.

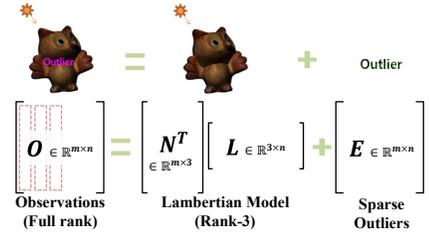

Fig. 16: Photometric stereo illustration for the used model.

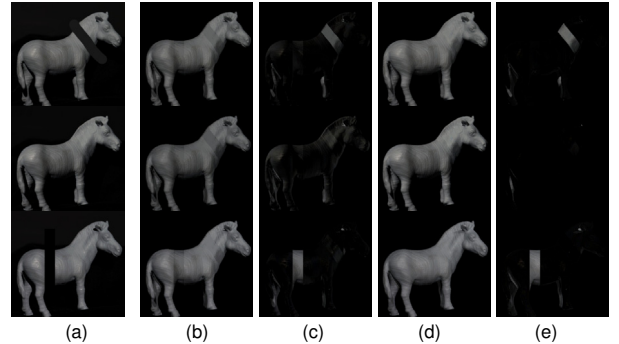

Fig. 17: Outlier rejection results for Photometric Stereo. (a) Three sampled input images out of five. (b-c) low-rank and sparse image from Wu *et al*. [45]. (d-e) low-rank and sparse image from ours.

### 4.2.2 Motion Detection by Temporal Edge

RPCA-based background modeling for surveillance purpose requires a large number of observations to estimate background and moving objects under global illumination changes. Such requirement is not suitable for online algorithm in surveillance. Using a few images as input, the moving region detection by RPCA could fail due to the limited number of observations. In this problem, we observe that edge images make moving object boundaries more sparse and they rarely overlap. We stack a few $n$ edge images (obtained by Sobel operator) in video sequence as column vectors of a matrix $\mathbf{O} \in \mathbb{R}^{m \times n} = [\text{vec}(O_1)| \cdots |\text{vec}(O_n)]$. Without moving objects, the edge pixels on the background texture are static, so the matrix $\mathbf{O}$ should be low-rank, essentially, rank-1. Since moving object regions are not consistent with background edges, the regions can be modeled as sparse outliers.

Fig. 15 shows the comparisons with RPCA and the proposed method. RPCA fails to decompose low–rank and sparse matrix in Fig. 15-(b) due to deficient observations where $n = 5$. On the other hand, our method successfully estimates moving object boundary, and the results are similar to the one obtained with many observations in Fig. 15-(e).

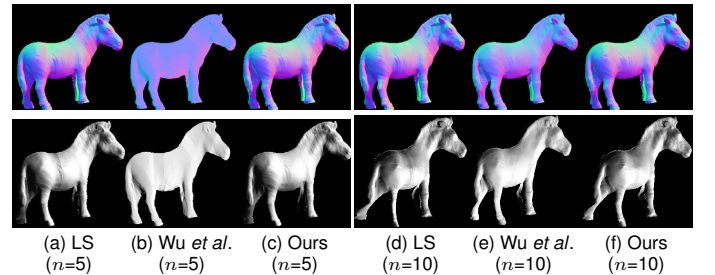

(a) LS ($n$=5)   (b) Wu *et al*. ($n$=5)   (c) Ours ($n$=5)   (d) LS ($n$=10)   (e) Wu *et al*. ($n$=10)   (f) Ours ($n$=10)

Fig. 18: Results of the photometric normal estimation and depth by LS method [42], Wu *et al*. [45] and our method. Top row: the normal estimation. Bottom row: the reconstructed depth from the estimated normals. $n$ denotes the number of input images.

|  | $\sigma_1$ | $\sigma_2$ | $\sigma_3$ | $\sigma_4$ | $\sigma_5$ |
|---|---|---|---|---|---|
| Input $\sigma_i(\mathbf{O}_G)$ | 137.88 | 23.59 | 19.55 | 15.56 | 12.99 |
| RPCA $\sigma_i(\hat{\mathbf{A}}_G)$ in Fig. 18-(b) | 125.71 | 7.26 | *0.0001* | 0.00 | 0.00 |
| Ours $\sigma_i(\hat{\mathbf{A}}_G)$ in Fig. 18-(c) | 139.16 | 23.01 | 16.33 | 1.59 | 0.15 |

TABLE 2: Singular values of photometric stereo input for $n=5$ in Fig. 17 and Fig. 18-(b,c).

### 4.2.3 Outlier Rejection for Photometric Stereo

Intensity observation is modeled in Lambertian photometric stereo as $\mathbf{O} = [\text{vec}(O_1)| \cdots |\text{vec}(O_n)] = \mathbf{N}^\top \mathbf{L}$[4], where $\mathbf{O} \in \mathbb{R}^{m \times n}$, $\mathbf{N} \in \mathbb{R}^{3 \times m}$ and $\mathbf{L} \in \mathbb{R}^{3 \times n}$ denote measured intensity, normal and light direction matrix, respectively, and $m$ and $n$ are the number of pixels and images. Hayakawa *et al.* [23] show that the intensity matrix lies in a subspace of rank 3, as illustrated in Fig. 16. However, this constraint is hardly satisfied in real situations due to shadow from self-occlusion, saturation and some object materials which do not exactly follow the Lambertian diffusion model. Considering the rank-3 constraint, the artifacts mentioned above can be regarded as sparse outliers and we get a low-rank structure as $\mathbf{O} = \mathbf{N}^\top \mathbf{L} + \mathbf{E}$.

The robust photometric stereo with outlier rejection can be formulated as a RPCA problem as suggested by Wu *et al.* [45]. We compare our method with the standard least square (LS) method [42] and RPCA by Wu *et al.* [45]. Among them, Wu *et al.* and our method do not require light information (i.e. uncalibrated setting), while the LS method requires light calibration a priori. Thus, RPCA based outlier rejection is a more challenging problem than the robust regression given light information [26]. The LS based photometric stereo estimates the normals by minimizing $\|\mathbf{O} - \mathbf{N}^\top \mathbf{L}\|_F^2$. We corrupt some input images by painted artifacts to mimic outliers. The corrupted inputs are included in 2 out of $n=5$ inputs (Fig. 17 and Fig. 18-(top)), and 4 out of $n=10$ inputs (Fig. 18-(bottom)). Outlier rejection results are shown in Fig. 17. We present qualitative comparison of normal recovery results in Fig. 17 and Fig. 18. Wu *et al.* return a planar surface normal when the rank of input matrix is lower than 3 due to the lack of observations (italic in Table 2). When more input images are available, RPCA begins to return detail preserved results, as shown in Fig. 18-(e). On the other hand, our method consistently provides robust results for both limited and sufficient observations, as shown in Fig. 18-(c,f).

For quantitative results, we use the *Bunny* dataset [26] generated using the Cook-Torrance reflectance model and consisting of 40 different lighting conditions. The average ratio of specular and shadow regions in *Bunny* are 8.4% and 24% respectively, which act as outliers. Table 3 shows quantitative results. We vary the number of images and add 5% of uniformly distributed corruption. Each value in Table 3 is averaged over 20 randomly selected test sets. Wu *et al.* [45] produce degenerated results, as the rank of the resulting matrix is lower than 3 due to the lack of supports from the observations. When more input images are available, RPCA returns more satisfying results, but still the accuracy is lower than LS method. In contrast,

---

4. Note that the intensities only on the object region are used in the observation matrix $\mathbf{O}$ with the corresponding object mask.

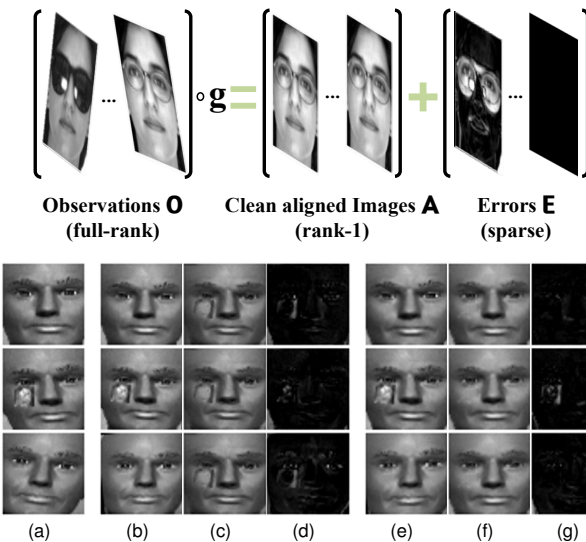

Observations **O** (full-rank) $\circ$ **g** = Clean aligned Images **A** (rank-1) + Errors **E** (sparse)

(a) (b) (c) (d) (e) (f) (g)

Fig. 20: Batch image alignment experiments. (a) Three input images. (b-d) The aligned, low-rank and sparse results by Peng *et al.* [38]. (e-g) The aligned, low-rank and sparse results by the proposed method. The images in the illustration on the top row are from AR dataset [32].

our method provides accurate results for both limited and sufficient observations.

### 4.2.4 Batch Image Alignment

Given several images of an object of interest (e.g. face), the batch image alignment task aims to align them to a fixed canonical template [6], [38]. In this problem, we search for a transformation $g_i$ for each image $I_i$ to make the images linearly correlated. We note **g** the set of transformations: $\mathbf{g} = \{g_1, \ldots, g_n\}$ where $n$ is the number of images and write $\mathbf{O} \circ \mathbf{g} = [\text{vec}(I_1 \circ g_i)| \cdots |\text{vec}(I_n \circ g_n)]$. Contrary to the formulation of Peng *et al.* [38], we consider PSSV mathematically formulated as follows:

$$\arg\min_{\mathbf{A},\mathbf{E},\mathbf{g}} \|\mathbf{A}\|_{p=N} + \lambda\|\mathbf{E}\|_1, \text{ s.t. } \mathbf{O} \circ \mathbf{g} = \mathbf{A} + \mathbf{E}. \quad (24)$$

We applied our approach to the head dataset acquired under varying poses (see Fig. 20-(a)) [38]. For linearly correlated noise-free batch images, the rank is $N=1$, when the transformations for exact image alignment are estimated. Our results of alignment, low-rank estimation and error sparsity are shown in Fig. 20-(e,f,g). Compared to the results obtained by RASL [38], our method can correctly detect the outliers (Fig. 20-(c) v.s. Fig. 20-(f)), even with only 3 input images.

Our method can correctly detect the outliers and also robustly align the images even if the geometric model has more degrees of freedom than an affine homography model. Detailed comparisons in Fig. 22 show the average image obtained from the aligned image stack by each method. If well aligned, the average image should show seamless image without duplicated edges. Our results show fine average images due to more accurate homography estimation than RASL.

### 4.2.5 Image Recovery

Images of natural scenes follow natural statistics [25]. As shown by Hu *et al.* [24], information of image scenes is dominated by the top 20 singular values, which is low-rank. Hu *et al.* [24] proposed a matrix completion method with the truncated nuclear norm (TNN) as introduced in Sec. 2. We formulate the matrix completion as

$$\arg\min_{\mathbf{A},\mathbf{B}} \|\mathbf{A}\|_p, \quad \text{s.t.} \quad \mathbf{A} = \mathbf{B}, \mathcal{P}_\Omega(\mathbf{B}) = \mathcal{P}_\Omega(\mathbf{O}), \quad (25)$$

where $\mathcal{P}_\Omega(\cdot)$ is the orthogonal projection operator setting $[\mathcal{P}_\Omega(\mathbf{X})]_{i,j} = [\mathbf{X}]_{i,j}$ for $(i,j) \in \Omega$ and 0 otherwise. Although the auxiliary variable **B** looks unnecessary, with the affine constraint (the projection operator) it makes the efficient PSVT operator applicable in ADMM algorithm. We set $\rho = 1.05$ in ADMM (see the supplementary material and Alg. 2), and for fairness, we follow the same setting as suggested by Hu *et al.*, i.e. $\mu_0 = 1e^{-3}$.

| | Mean error (in degrees) | | | Max error (in degrees) | | | Standard deviation | | |
|---|---|---|---|---|---|---|---|---|---|
| No. Image | LS [42] | Wu et al. [45] | Ours | LS [42] | Wu et al. [45] | Ours | LS [42] | Wu et al. [45] | Ours |
| 5 | 8.53 | 27.88 | **7.06** | 159.72 | 130.77 | **120.78** | 14.48 | 16.45 | **12.30** |
| 8 | 9.03 | 13.34 | **5.87** | 142.45 | 139.07 | **85.48** | 11.24 | 10.96 | **9.62** |
| 10 | 9.24 | 11.14 | **5.70** | 148.05 | 110.12 | **79.54** | 9.91 | 9.77 | **8.06** |
| 12 | 8.96 | 9.95 | **5.09** | 130.04 | 80.21 | **76.86** | 9.17 | 9.02 | **7.59** |

TABLE 3: Photometric stereo results of *Bunny* with 5% corruption ratio, additional specularities and shadows. Bold fonts indicate highest accuracy.

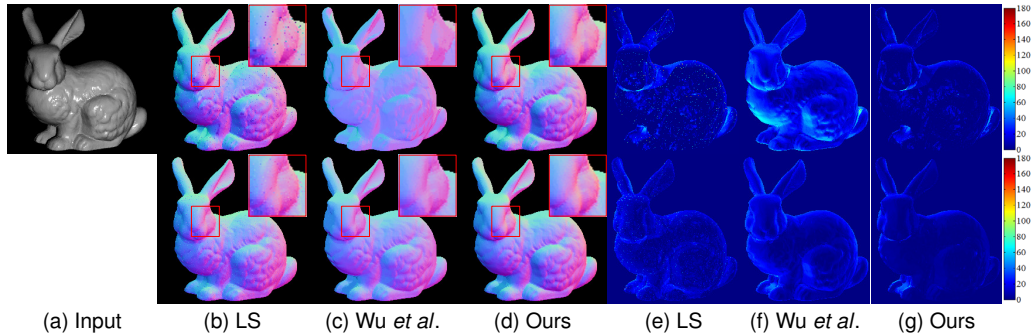

(a) Input (b) LS (c) Wu *et al.* (d) Ours (e) LS (f) Wu *et al.* (g) Ours

Fig. 19: Photometric stereo results from 5 (top) and 12 (bottom) images of *Bunny* dataset with corruption. (a) A representative input image. (b-d) Recovered surface normals by LS [42], Wu *et al*. [45] and ours. (e-g) Corresponding error maps for each algorithm.

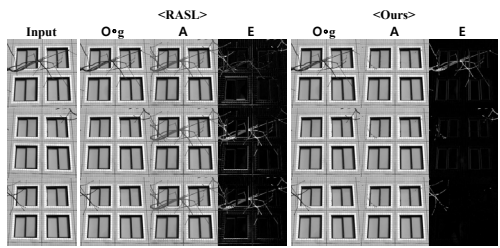

Fig. 21: Aligning planar surfaces despite occlusions by RASL [38] and ours with $n = 4$ images. Affine transformation is used as the geometric transformation model **g**.

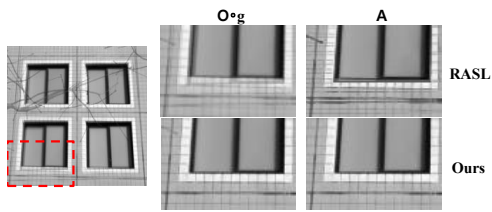

Fig. 22: Close-up comparisons of Fig. 21 (see red box in the left) with the average images of aligned results **O**∘**g** and recovered low-rank components **A** between RASL [38] and ours.

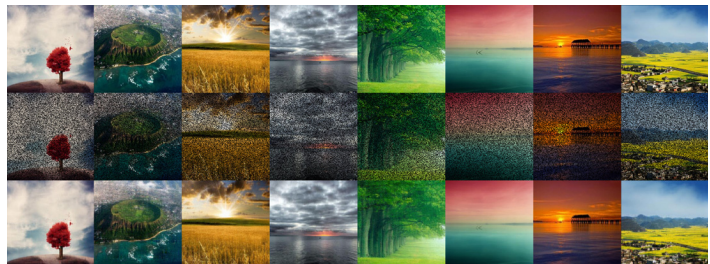

| Index | | 1 | 2 | 3 | 4 | 5 | 6 | 7 | 8 |
|---|---|---|---|---|---|---|---|---|---|
| PSNR | [24] | 30.428 | 21.741 | 24.506 | 32.282 | 24.375 | 42.145 | 33.357 | 21.293 |
| | Ours | 30.429 | 21.768 | 24.507 | 32.216 | 24.376 | 42.011 | 33.282 | 21.325 |
| Iter. | [24] | 755 | 727 | 631 | 761 | 639 | 678 | 1251 | 799 |
| | Ours | 183 | 210 | 201 | 196 | 192 | 181 | 216 | 201 |
| Time (sec) | [24] | 18.60 | 18.36 | 15.49 | 18.77 | 15.87 | 16.59 | 20.84 | 19.35 |
| | Ours | 4.06 | 4.30 | 4.48 | 4.20 | 4.09 | 3.94 | 3.07 | 4.47 |

Fig. 23: Image recovery application. Top: Original images (ground truth). Middle: Observed images with 50% missing entries (input). Bottom: Recovered image by our method. Table: Quantitative comparison with Hu *et al*. [24] for PSNR, number of iterations, and running time.

Fig. 23 shows the comparison with Hu *et al.* Since the proposed method has a similar objective function with Hu *et al.*, PSNR of recovered images are similar (the max. difference is 0.134), but our method requires fewer iterations and runs 4 times faster. Because Hu *et al.* perform local approximation to solve PSSV, their algorithm requires outer loop to fix subspaces and inner loop to minimize nuclear norm with affine constraint. In contrast, our method directly minimizes PSSV, which is a key contribution of our work.

## 5 DISCUSSIONS AND CONCLUSION

In this paper, we revisited the rank minimization method in RPCA for low-level vision problems. When the target rank is known, we show that, by modifying the objective function from the nuclear norm to PSSV, we can achieve a better control of the target rank of the low-rank solution, even when the number of observations is limited. The appealing advantage of our solution is that it can be easily utilized in existing algorithms, e.g. ADMM [31], and the efficient computation properties still hold. The generality and the effectiveness of our approach are supported through numerous and extensive experiments on both synthetic examples and several real-world applications which outperform the conventional nuclear norm objective function. We do not consider scalability issues of our method in this paper, but the recent approach suggested by Oh *et al.* [37] allow to speed-up the application of our method. An interesting direction of future work is the mathematical analysis of the properties of our partial sum objective function compared to the nuclear norm solution. In the following, we discuss some open questions related to our paper.

**Sufficient number of samples versus minimum number of samples** In our experimental analysis, we found that our solution is more robust than the nuclear norm solution when facing a limited number of samples. Defining $K$ as the (theoretical) minimum number of samples for processing, e.g. 2 images for HDR, 3 images for photometric stereo, our approach requires more than $K$ samples for a robust model estimation and outlier rejection. We believe that the number of needed additional samples depends on the problem setting, e.g. the shape of feature space or the distribution of the samples.

**Target rank** While our formulation implicitly encourages a target rank constraint in the resulting matrix, this constraint is hardly enforced. We discuss here two possible scenarios that can produce the resulting matrix having a rank different from the target rank. The first scenario is when a very limited number of samples are observed. In such case, PSVT can produce a deficient rank lower than the target rank when the span of the observed samples is less than the target rank, but this case is a fundamental limitation of under-sampling rather than a conceptual limitation of our approach. Another scenario is due to too much noise (especially for Gaussian noise that does not follow the sparsity property) in the observed samples which results in large singular values in the residual ranks. In this case, a solution to satisfy the rank constraint is to increase $\tau$ in Eq. (17). When $\tau$ is equal to infinity, our PSVT solution is close to the result using singular value projection [27]. However, the projection method enforcing target rank could produce an over-fitting solution due to the mentioned noise effects.

## ACKNOWLEDGMENTS

We would like to thank reviewers and associate editor for their valuable comments, and thank Steve Seitz and Dan Goldman for the photometric stereo dataset. The work was supported by the National Research Foundation of Korea (NRF) grant funded by the Korea government (MSIP) (No. 2010-0028680). In So Kweon is the corresponding author.

## REFERENCES

[1] A. Agarwal, S. Negahban, and M. J. Wainwright. Noisy matrix decomposition via convex relaxation: Optimal rates in high dimensions. In *International Conference on Machine Learning (ICML)*, 2011.
[2] F. Bach, J. Mairal, and J. Ponce. Convex sparse matrix factorizations. *arXiv preprint arXiv:0812.1869*, 2008.
[3] R. Basri, D. Jacobs, and I. Kemelmacher. Photometric stereo with general, unknown lighting. *International Journal of Computer Vision (IJCV)*, 72(3):239–257, 2007.
[4] S. Boyd, N. Parikh, E. Chu, B. Peleato, and J. Eckstein. Distributed optimization and statistical learning via the alternating direction method of multipliers. *Foundations and Trends® in Machine Learning*, 3(1):1–122, 2011.
[5] C. Bregler, A. Hertzmann, and H. Biermann. Recovering non-rigid 3D shape from image streams. In *IEEE Conference on Computer Vision and Pattern Recognition (CVPR)*, 2000.
[6] L. G. Brown. A survey of image registration techniques. *ACM Computing Surveys*, 24(4):325–376, 1992.
[7] R. Cabral, F. D. L. Torre, J. P. Costeira, and A. Bernardino. Unifying nuclear norm and bilinear factorization approaches for low-rank matrix decomposition. In *IEEE International Conference on Computer Vision (ICCV)*, 2013.
[8] J.-F. Cai, E. J. Candès, and Z. Shen. A singular value thresholding algorithm for matrix completion. *SIAM Journal on Optimization*, 20(4):1956–1982, 2010.
[9] E. Candès, X. Li, Y. Ma, and J. Wright. Robust principal component analysis? *Journal of the ACM*, 58(3):11, 2011.
[10] V. Chandrasekaran, S. Sanghavi, P. A. Parrilo, and A. S. Willsky. Rank-sparsity incoherence for matrix decomposition. *SIAM Journal on Optimization*, 21(2):572–596, 2011.
[11] K. Chen, H. Dong, and K.-S. Chan. Reduced rank regression via adaptive nuclear norm penalization. *Biometrika*, 2013.
[12] F. De la Torre and M. Black. Robust principal component analysis for computer vision. In *IEEE International Conference on Computer Vision (ICCV)*, 2001.
[13] F. De la Torre and M. Black. A framework for robust subspace learning. *International Journal of Computer Vision (IJCV)*, 54(1-3):117–142, 2003.
[14] E. M. de Sá. Exposed faces and duality for symmetric and unitarily invariant norms. *Linear Algebra and its Applications*, 197:429–450, 1994.
[15] P. Debevec and J. Malik. Recovering high dynamic range radiance maps from photographs. In *Proceedings of the 24th annual conference on Computer graphics and interactive techniques*, pages 369–378, 1997.
[16] D. L. Donoho. High-dimensional data analysis: the curses and blessings of dimensionality. In *American Mathematical Society Conf. Math Challenges of the 21st Century*, pages 1–32, 2000.
[17] D. L. Donoho and I. M. Johnstone. Adapting to unknown smoothness via wavelet shrinkage. *Journal of the American Statistical Association*, 90(432):1200–1224, 1995.
[18] A. Eriksson and A. van den Hengel. Efficient computation of robust weighted low-rank matrix approximations using the $l_1$ norm. *IEEE Transactions on Pattern Analysis and Machine Intelligence (TPAMI)*, 34(9):1681–1690, 2012.
[19] M. Fischler and R. Bolles. Random sample consensus: A paradigm for model fitting with applications to image analysis and automated cartography. *Communications of the ACM*, 24(6):381–395, 1981.
[20] S. Gaïffas and G. Lecué. Weighted algorithms for compressed sensing and matrix completion. *arXiv preprint arXiv:1107.1638*, 2011.
[21] O. Gallo, N. Gelfand, W. Chen, M. Tico, and K. Pulli. Artifact-free high dynamic range imaging. *IEEE International Conference on Computational Photography (ICCP)*, 2009.
[22] E. T. Hale, W. Yin, and Y. Zhang. Fixed-point continuation for $l_1$-minimization: Methodology and convergence. *SIAM Journal on Optimization*, 19(3):1107–1130, 2008.
[23] H. Hayakawa. Photometric stereo under a light source with arbitrary motion. *Journal of the Optical Society of America A*, 11(11):3079–3089, 1994.
[24] Y. Hu, D. Zhang, J. Ye, X. Li, and X. He. Fast and accurate matrix completion via truncated nuclear norm regularization. *IEEE Transactions on Pattern Analysis and Machine Intelligence (TPAMI)*, 35(9):2117–2130, 2013.
[25] A. Hyvärinen, J. Hurri, and P. O. Hoyer. *Natural Image Statistics: A Probabilistic Approach to Early Computational Vision*, volume 39. Springer, 2009.
[26] S. Ikehata, D. Wipf, Y. Matsushita, and K. Aizawa. Robust photometric stereo using sparse regression. In *IEEE Conference on Computer Vision and Pattern Recognition (CVPR)*, 2012.
[27] P. Jain, R. Meka, and I. Dhillon. Guaranteed rank minimization via singular value projection. In *Advances in Neural Information Processing Systems (NIPS)*, 2010.
[28] H. Ji, C. Liu, Z. Shen, and Y. Xu. Robust video denoising using low rank matrix completion. In *IEEE Conference on Computer Vision and Pattern Recognition (CVPR)*, 2010.
[29] I. T. Jolliffe. Principal Component Analysis. In *Springer-Verlag*, 1986.
[30] Q. Ke and T. Kanade. Robust $l_1$ norm factorization in the presence of outliers and missing data by alternative convex programming. In *IEEE Conference on Computer Vision and Pattern Recognition (CVPR)*, 2005.
[31] Z. Lin, M. Chen, and Y. Ma. The augmented Lagrange multiplier method for exact recovery of corrupted low-rank matrices. Technical Report UILU-ENG-09-2215, UIUC, 2009.
[32] A. Martinez and R. Benavente. The AR face database. *CVC Technical Report no. 24*, 1998.
[33] T.-H. Oh, H. Kim, Y.-W. Tai, J.-C. Bazin, and I. S. Kweon. Partial sum minimization of singular values in RPCA for low-level vision. In *IEEE International Conference on Computer Vision (ICCV)*, 2013.
[34] T.-H. Oh, J.-Y. Lee, and I. S. Kweon. Real-time motion detection based on discrete cosine transform. In *IEEE International Conference on Image Processing (ICIP)*, 2012.
[35] T.-H. Oh, J.-Y. Lee, and I. S. Kweon. High dynamic range imaging by a rank-1 constraint. In *IEEE International Conference on Image Processing (ICIP)*, 2013.
[36] T.-H. Oh, J.-Y. Lee, Y.-W. Tai, and I. S. Kweon. Robust high dynamic range imaging by rank minimization. *IEEE Transactions on Pattern Analysis and Machine Intelligence (TPAMI)*, 37(6):1219–1232, 2015.

[37] T.-H. Oh, Y. Matsushita, Y.-W. Tai, and I. S. Kweon. Fast randomized singular value thresholding for nuclear norm minimization. In *IEEE Conference on Computer Vision and Pattern Recognition (CVPR)*, 2015.
[38] Y. Peng, A. Ganesh, J. Wright, W. Xu, and Y. Ma. RASL: Robust alignment by sparse and low-rank decomposition for linearly correlated images. *IEEE Transactions on Pattern Analysis and Machine Intelligence (TPAMI)*, 34(11):2233–2246, 2012.
[39] B. Recht, W. Xu, and B. Hassibi. Necessary and sufficient conditions for success of the nuclear norm heuristic for rank minimization. In *IEEE Conference on Decision and Control*, 2008.
[40] Y. Shen, Z. Wen, and Y. Zhang. Augmented lagrangian alternating direction method for matrix separation based on low-rank factorization. *Optimization Methods and Software*, 29(2):239–263, 2014.
[41] C. Tomasi and T. Kanade. Shape and motion from image streams under orthography: a factorization method. *International Journal of Computer Vision (IJCV)*, 9(2):137–154, 1992.
[42] R. Woodham. Photometric method for determining surface orientation from multiple images. In *Optical Engineering*, 1980.
[43] J. Wright, A. Ganesh, S. Rao, and Y. Ma. Robust principal component analysis: Exact recovery of corrupted low-rank matrices via convex optimization. In *Advances in Neural Information Processing Systems (NIPS)*, 2009.
[44] J. Wright, A. Ganesh, Z. Zhou, K. Min, S. Rao, Z. Lin, Y. Peng, M. Chen, L. Wu, Y. Ma, E. Candès, and X. Li. Low-rank matrix recovery and completion via convex optimization. http://perception.csl.illinois.edu/matrix-rank/home.html.
[45] L. Wu, A. Ganesh, B. Shi, Y. Matsushita, Y. Wang, and Y. Ma. Robust photometric stereo via low-rank matrix completion and recovery. In *Asian Conference on Computer Vision (ACCV)*, 2010.
[46] Y. Wu and D. P. Wipf. Dual-space analysis of the sparse linear model. In *Advances in Neural Information Processing Systems (NIPS)*, pages 1745–1753, 2012.
[47] Z. Zhang, A. Ganesh, X. Liang, and Y. Ma. TILT: Transform invariant low-rank textures. *International Journal of Computer Vision (IJCV)*, 99(1):1–24, 2012.
[48] Y. Zheng, G. Liu, S. Sugimoto, S. Yan, and M. Okutomi. Practical low-rank matrix approximation under robust $l_1$-norm. In *IEEE Conference on Computer Vision and Pattern Recognition (CVPR)*, 2012.
[49] Z. Zhou, X. Li, J. Wright, E. Candes, and Y. Ma. Stable principal component pursuit. In *IEEE International Symposium on Information Theory Proceedings (ISIT)*. IEEE, 2010.

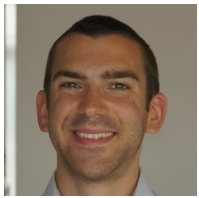

**Jean-Charles Bazin** is an Associate Research Scientist at Disney Research Zurich, Switzerland. Before joining Disney Research, he was a postdoc at the Computer Graphics Laboratory of Prof. Markus Gross and at the Computer Vision and Geometry Group of Prof. Marc Pollefeys at ETH Zurich, Switzerland (2011-2014). He was also a Postdoctoral Fellow at Computer Vision Lab of Prof. Katsushi Ikeuchi, Tokyo, Japan (2010/2011). He obtained his MS in Computer Science at Universite de Technologie de Compiegne, France (2006) and his PhD in Electrical Engineering from KAIST, South Korea (2011).

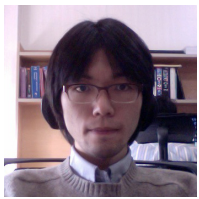

**Hyeongwoo Kim** received his Bachelor degree in Electrical Engineering and Computer Science from Yonsei University in 2005 and his Master degree in Electrical Engineering and Computer Science from KAIST, South Korea in 2007. His research interests include computer vision and image processing.

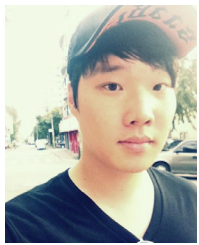

**Tae-Hyun Oh** received the B.E degree (summa cum laude) in Computer Engineering from Kwang-Woon University in 2010, and the M.S degree in Electrical Engineering from KAIST, South Korea, in 2012. He is currently working towards the Ph.D. degree at KAIST. He was a visiting student in the Visual Computing Group, Microsoft Research Asia. He was a recipient of Gold prize of Samsung HumanTech Thesis Award and Qualcomm Innovation Award. His research interests include robust computer vision and machine learning. He is a student member of the IEEE.

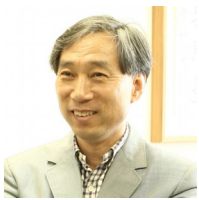

**In So Kweon** received the BS and MS degrees in mechanical design and production engineering from Seoul National University, in 1981 and 1983, respectively, and the PhD degree in robotics from Carnegie Mellon University, 1990. He is now a professor at KAIST, South Korea. He is a recipient of the best student paper runner-up award at CVPR 09'. He was the program chair for ACCV 07' and is the general chair for ACCV 12'. He is also on the editorial board of the IJCV.

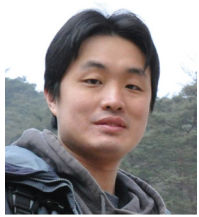

**Yu-Wing Tai** received the Ph.D. degree in Computer Science from the National University of Singapore (NUS) in 2009, the M.S. degree and the Bachelor degree in Computer Science from the Hong Kong University of Science and Technology (HKUST) in 2005 and 2003 respectively. He is now an associate professor in EE department of KAIST, South Korea. He regularly serves on the program committees for the major Computer Vision conferences (ICCV, CVPR and ECCV). His research interests include computer vision and image/video processing. He is a senior member of the IEEE.

# Supplementary Material: Partial Sum Minimization of Singular Values in Robust PCA: Algorithm and Applications

Tae-Hyun Oh *Student Member, IEEE,* Yu-Wing Tai *Senior Member, IEEE,*
Jean-Charles Bazin *Member, IEEE,* Hyeongwoo Kim *Student Member, IEEE,*
and In So Kweon *Member, IEEE*

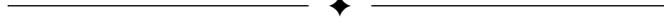

## SUPPLEMENTARY MATERIAL

This is the supplementary material for the main paper [5]. In this supplementary material, we prove Proposition 1, and we provide the pseudo code of the algorithm of image recovery application. We also present an additional experimental result not included in the main paper due to space limitation. All the parameters are the same as in the main paper or the referred papers, except if stated otherwise.

## 1 PROOFS OF PROPOSITION 1.

**Lemma 4** (Lipschitz continuous of PSSV). *The function of the partial sum of singular values $h(\mathbf{X}) = \|\mathbf{X}\|_p = \sum_{i=p+1}^{\min(m,n)} \sigma_i(\mathbf{X})$ (where $p \in \mathbb{N}$ denotes the target rank) for $\mathbf{X} \in \mathbb{R}^{m \times n}$ is Lipschitz continuous. Namely, there exists a constant scalar $K$ satisfying*

$$|h(\mathbf{X}_1) - h(\mathbf{X}_2)| \leq K \cdot \|\mathbf{X}_1 - \mathbf{X}_2\|_F \quad \text{for all } \mathbf{X}_1, \mathbf{X}_2 \in \mathbb{R}^{m \times n}.$$

*Proof.* Let the nuclear norm be $f(\mathbf{X}) = \|\mathbf{X}\|_* = \sum_{i=1}^{\min(m,n)} \sigma_i(\mathbf{X})$, and the Ky-Fan $p$-norm be $g(\mathbf{X}) = \|\mathbf{X}\|_{Ky(p)} = \sum_{i=1}^{p} \sigma_i(\mathbf{X})$. By definition, $h(\mathbf{X}) = \|\mathbf{X}\|_p = f(\mathbf{X}) - g(\mathbf{X})$, and we know that the nuclear norm $f(\cdot)$ [4] and the Ky-Fan $p$-norm $g(\cdot)$ [6] are Lipschitz continuous (derivation of the Lipschitz continuity for Ky-Fan matrix norm is straight forward). Therefore we have

$$\begin{aligned}|f(\mathbf{X}_1) - f(\mathbf{X}_2)| &\leq K_f \cdot \|\mathbf{X}_1 - \mathbf{X}_2\|_F, \\ |g(\mathbf{X}_1) - g(\mathbf{X}_2)| &\leq K_g \cdot \|\mathbf{X}_1 - \mathbf{X}_2\|_F,\end{aligned} \quad (1)$$

where $K_f$ and $K_g$ are Lipschitz constants for $f$ and $g$ respectively.

We see that
$$\begin{aligned}|h(\mathbf{X}_1) - h(\mathbf{X}_2)| =& |f(\mathbf{X}_1) - g(\mathbf{X}_1) - (f(\mathbf{X}_2) - g(\mathbf{X}_2))| \\ =& |f(\mathbf{X}_1) - f(\mathbf{X}_2) + (g(\mathbf{X}_2) - g(\mathbf{X}_1))| \\ \leq& |f(\mathbf{X}_1) - f(\mathbf{X}_2)| + |g(\mathbf{X}_1) - g(\mathbf{X}_2)| \quad \text{(by triangle inequality)} \\ \leq& (K_f + K_g) \cdot \|\mathbf{X}_1 - \mathbf{X}_2\|_F.\end{aligned}$$

Since the constant $K = K_f + K_g$ satisfies the inequality, $h(\mathbf{X}) = \|\mathbf{X}\|_p$ is Lipschitz continuous. $\square$

Since PSSV $\|\cdot\|_p$ is a non-convex function, the typical subdifferential for convex functions (a.k.a. Fenchel-Moreau subdifferential for convex functions. Refer to Daniilidis *et al.* [3]) would be the empty set. Therefore, we introduce a generalized subdifferential (a.k.a. Clarke subdifferential, see Definition 3.2 in [1]) for non-convex locally Lipschitz continuous functions. This is useful for deriving stationary points in the convergence proof.

**Definition 2** (Generalized subgradients). *Let $f : \mathbb{R}^n \to \mathbb{R}$ be a locally Lipschitz continuous function at a point $\mathbf{x} \in \mathbb{R}^n$. Then the **subdifferential** of $f$ at $\mathbf{x}$ is the set $\partial_C f(\mathbf{x})$ of vectors $\mathbf{z} \in \mathbb{R}^n$ such that*

$$\partial_C f(\mathbf{x}) = \{\mathbf{z} : f^\circ(\mathbf{x}; \mathbf{d}) \geq \langle \mathbf{z}, \mathbf{d} \rangle \text{ for all } \mathbf{d} \in \mathbb{R}^n\}, \quad (2)$$

*where each vector $\mathbf{z} \in \partial_C f(\mathbf{x})$ is called a subgradient of $f$ at $\mathbf{x}$, and the directional subgradient of $f$ at $\mathbf{x}$ in the direction vector $\mathbf{d} \in \mathbb{R}^n$ is defined as $f^\circ(\mathbf{x}; \mathbf{d}) = \limsup\limits_{\substack{\mathbf{y} \to \mathbf{x} \\ t \downarrow 0}} \frac{f(\mathbf{y} + t\mathbf{d}) - f(\mathbf{y})}{t}$.*

**Remark D.2.1.** *Definition 2 can be generalized to matrix cases analogously.*

**Remark D.2.2.** *Regardless of non-convexity or non-smoothness, the generalized subdifferential (here, Clarke subdifferential) always exists for locally Lipschitz continuous functions. Also, $\partial_C \|\cdot\|_p$ is well defined in $\mathbb{R}^{m\times n}$, because $\|\cdot\|_p$ is a Lipschitz continuous function as shown in Lemma 4.*

The following Lemma is also important for the convergence proof of Proposition 1.

**Lemma 5** (Convexity and compactness of subdifferential [1]) *Let $f: \mathbb{R}^{m\times n} \to \mathbb{R}$ be a locally Lipschitz continuous function at $\mathbf{X}$. Then the subdifferential $\partial_C f(\mathbf{X})$ is a non-empty, convex and compact set.*

**Remark L.5.1.** *Basic properties of the generalized subdifferential are identical to those in convex case, and most of subdifferential calculus rules hold. Note that* Karush-Kuhn-Tucker (KKT) optimality conditions *are also properly defined with the generalized subdifferential [1].*

Since our problem in Eq. (3, Main) does not have any inequality constraint, but an equality constraint, the KKT conditions are reduced to stationary and primal feasibility conditions. Armed with the above lemmas and definitions, we can now propose and prove the convergence of Alg. 1.

**Proposition 1** (Convergence). *Let $S_k = (\mathbf{A}_k, \mathbf{E}_k, \mathbf{Y}_k, \hat{\mathbf{Y}}_k)$, where $\hat{\mathbf{Y}}_{k+1} = \mathbf{Y}_k + \mu_k(\mathbf{O} - \mathbf{A}_{k+1} - \mathbf{E}_k)$ and $\{S_k\}_{k=1}^\infty$ is a set of intermediate solutions of Alg. 1. Suppose that $\{\mathbf{Y}_k\}_{k=1}^\infty$ and $\{\hat{\mathbf{Y}}_k\}_{k=1}^\infty$ are bounded, $\lim_{k\to\infty}(\mathbf{Y}_{k+1} - \mathbf{Y}_k) = 0$, and $\mu_k$ is non-decreasing, then any accumulation point of $\{S_k\}_{k=1}^\infty$ satisfies the following KKT conditions:*

$$(C1)\ \mathbf{Y}^* \in \partial_C\|\mathbf{A}^*\|_p, \quad (C2)\ \mathbf{Y}^* \in \partial\|\lambda\mathbf{E}^*\|_1, \quad (C3)\ \mathbf{O} - \mathbf{A}^* - \mathbf{E}^* = \mathbf{0}, \quad (C4)\ \partial_C\|\mathbf{A}^*\|_p \cap \partial\|\lambda\mathbf{E}^*\|_1 \neq \emptyset, \quad (3)$$

*where $\mathbf{Y}^*, \mathbf{A}^*$ and $\mathbf{E}^*$ represent each cluster points. In particular, whenever $\{S_k\}_{k=1}^\infty$ converges, it converges to a KKT point of Eq. (2, Main).*

*Proof.* For $\mathbf{Y}$, we have $\mu_k^{-1}(\mathbf{Y}_{k+1} - \mathbf{Y}_k) = \mathbf{O} - \mathbf{A}_{k+1} - \mathbf{E}_{k+1}$. Since $\lim_{k\to\infty}(\mathbf{Y}_{k+1} - \mathbf{Y}_k) = 0$ and $\mu_k$ is non-decreasing, we have $\mathbf{O} - \mathbf{A}_{k+1} - \mathbf{E}_{k+1} = \mu_k^{-1}(\mathbf{Y}_{k+1} - \mathbf{Y}_k) \to \mathbf{0}$, which satisfies (C3).

Since $\mathbf{E}_{k+1}$ obtained by the soft-thresholding operator [2] minimizes $L_{\mu_k}(\mathbf{A}_{k+1}, \mathbf{E}, \mathbf{Y}_k)$ (refer to Eq. (10, Main)) by definition, we have

$$\begin{aligned}\mathbf{0} \in &\partial\|\lambda\mathbf{E}_{k+1}\|_1 - \mathbf{Y}_k - \mu_k(\mathbf{O} - \mathbf{A}_{k+1} - \mathbf{E}_{k+1}) \\ =& \partial\|\lambda\mathbf{E}_{k+1}\|_1 - \mathbf{Y}_{k+1} \\ & \text{(since } \mathbf{Y}_{k+1} = \mathbf{Y}_k + \mu_k(\mathbf{O} - \mathbf{A}_{k+1} - \mathbf{E}_{k+1})\text{)} \\ \Rightarrow & \mathbf{Y}_{k+1} \in \partial\|\lambda\mathbf{E}_{k+1}\|_1,\end{aligned} \quad (4)$$

which satisfies (C2).

Since $\mathbf{A}_{k+1}$ optimally obtained by the PSVT minimizes $L_{\mu_k}(\mathbf{A}, \mathbf{E}_k, \mathbf{Y}_k)$ by Theorem 1 and Lemma 5, we have

$$\begin{aligned}\mathbf{0} \in & \partial_C\|\mathbf{A}_{k+1}\|_p - \mathbf{Y}_k - \mu_k(\mathbf{O} - \mathbf{A}_{k+1} - \mathbf{E}_k) \\ =& \partial_C\|\mathbf{A}_{k+1}\|_p - \mathbf{Y}_k - \mu_k(\mathbf{O} - \mathbf{A}_{k+1} - \mathbf{E}_{k+1}) - \mu_k(\mathbf{E}_{k+1} - \mathbf{E}_k) \\ =& \partial_C\|\mathbf{A}_{k+1}\|_p - \mathbf{Y}_{k+1} - \mu_k(\mathbf{E}_{k+1} - \mathbf{E}_k) \quad \text{(by definition of } \mathbf{Y}_{k+1}) \\ \Rightarrow & \mathbf{Y}_{k+1} + \mu_k(\mathbf{E}_{k+1} - \mathbf{E}_k) \in \partial_C\|\mathbf{A}_{k+1}\|_p.\end{aligned} \quad (5)$$

Since $\{\mathbf{Y}_k\}_{k=1}^\infty$ and $\{\hat{\mathbf{Y}}_k\}_{k=1}^\infty$ are bounded, there must exist a scalar $c > 0$ such that $\|\mathbf{Y}_{k+1}\|_F \leq c$ and $\|\hat{\mathbf{Y}}_{k+1}\|_F \leq c$. Then,

$$\begin{aligned}\mathbf{Y}_{k+1} - \hat{\mathbf{Y}}_{k+1} =& \mu_k\left(\mu_k^{-1}(\mathbf{Y}_{k+1} - \mathbf{Y}_k) - (\mathbf{O} - \mathbf{A}_{k+1} - \mathbf{E}_k)\right) \\ =& \mu_k((\mathbf{O} - \mathbf{A}_{k+1} - \mathbf{E}_{k+1}) - (\mathbf{O} - \mathbf{A}_{k+1} - \mathbf{E}_k)) = \mu_k(\mathbf{E}_k - \mathbf{E}_{k+1}) \\ \Rightarrow & \|\mathbf{E}_{k+1} - \mathbf{E}_k\|_F = \mu_k^{-1}\|\mathbf{Y}_{k+1} - \hat{\mathbf{Y}}_{k+1}\|_F \\ \leq & \mu_k^{-1}(\|\mathbf{Y}_{k+1}\|_F + \|\hat{\mathbf{Y}}_{k+1}\|_F) \quad \text{(by triangle inequality)} \\ \leq & 2c\mu_k^{-1} \to 0 \quad \text{(since } \mu_k \text{ is non-decreasing)}.\end{aligned} \quad (6)$$

Thus, $\mathbf{Y}_{k+1} \to \mathbf{Y}^* \in \partial_C\|\mathbf{A}^*\|_p$ from Eq. (5) and $\mathbf{Y}^* \in \partial_C\|\mathbf{A}^*\|_p \cap \partial\|\lambda\mathbf{E}^*\|_1 \neq \emptyset$ which satisfy (C1) and (C4). The sequence $\{S_k\}_{k=1}^\infty$ gradually satisfies the KKT conditions, which completes the proof. □

**Remark P.1.1.** *In Alg. 1 (and Proposition 1), the assumption for non-decreasing $\mu_k$ is always satisfied by the update rule $\mu_{k+1} = \rho\mu_k$ (with $\rho > 1$), and the boundness of the sequences $\{\mathbf{Y}_k\}_{k=1}^\infty$ and $\{\hat{\mathbf{Y}}_k\}_{k=1}^\infty$ is satisfied by the below Lemmas 6*

3*and 7 in conjunction with a similar manner of Lemma 1 in Lin* et al. *[4]. Therefore, we see that Alg. 1 converges as long as* $\mathbf{Y}_k$ *converges.*

**Lemma 6** (Boundness of $|f^\circ(\cdot;\cdot)|$ [1]) *Let $f : \mathbb{R}^{m\times n} \to \mathbb{R}$ be a locally Lipschitz continuous function at $\mathbf{X}$, and $\mathbf{d} \in \mathbb{R}^{m\times n}$. Then, there exists a scalar $B$ such that*

$$|f^\circ(\mathbf{X};\mathbf{d})| \leq B\|\mathbf{d}\|_F. \tag{7}$$

*Proof.* Refer to the proof of Theorem 3.1 in [1]. □

**Lemma 7** (Boundness of Clarke subgradient) *Let $f : \mathbb{R}^{m\times n} \to \mathbb{R}$ be a locally Lipschitz continuous function at $\mathbf{X}$. Then, a subgradient $\mathbf{W} \in \partial_C f(\mathbf{X})$ is bounded as*

$$\|\mathbf{W}\|_F \leq B, \tag{8}$$

*where $B$ is the same constant as in Lemma 6.*

*Proof.* By the definition of the Clarke subdifferential, $\mathbf{W}$ satisfies $f^\circ(\mathbf{X};\mathbf{d}) \geq \langle \mathbf{W}, \mathbf{d}\rangle$ for all $\mathbf{d} \in \mathbb{R}^{m\times n}$. By setting $\mathbf{d} = \mathbf{W}$, we get $f^\circ(\mathbf{X};\mathbf{W}) \geq \langle \mathbf{W},\mathbf{W}\rangle = \|\mathbf{W}\|_F^2$. Then, by Lemma 6, we have

$$\|\mathbf{W}\|_F^2 \leq |f^\circ(\mathbf{X};\mathbf{W})| \leq B\|\mathbf{W}\|_F \Rightarrow \|\mathbf{W}\|_F^2 \leq B\|\mathbf{W}\|_F$$
$$\Rightarrow \|\mathbf{W}\|_F \leq B \tag{9}$$

□

## 2 ALGORITHM FOR IMAGE RECOVERY

---

**Algorithm 2** ADMM for Image Recovery

---

**Input :** $\mathbf{O} \in \mathbb{R}^{m\times n}$, the index map $\Omega$, the constraint rank $N$.
Initialize $\mathbf{A}_0 = \mathbf{O}$, $\mathbf{B}_0 = \mathbf{Z}_0 = \mathbf{0}$, $\mu_0 > 0$, $\rho > 1$ and $k = 0$.
**while** not converged **do**
 $\mathbf{A}_{k+1} = \mathbb{P}_{N,\mu_k^{-1}}[\mathbf{B}_k - \mu_k^{-1}\mathbf{Z}_k]$.
 $\mathbf{B}_{k+1} = \underset{\mathcal{P}_\Omega(\mathbf{B})=\mathcal{P}_\Omega(\mathbf{O})}{\arg\min} \|\mathbf{B} - (\mathbf{A}_{k+1} + \mu_k^{-1}\mathbf{Z}_k)\|_F^2$.
 $\mathbf{Z}_{k+1} = \mathbf{Z}_k + \mu_k(\mathbf{A}_{k+1} - \mathbf{B}_{k+1})$.
 $\mu_{k+1} = \rho\mu_k$.
 $k = k + 1$.
**end while**
**Output :** $\mathbf{A}_k$.

---

## REFERENCES

[1] A. Bagirov, N. Karmitsa, and M. Makela. Nonconvex analysis. In *Introduction to Nonsmooth Optimization*, pages 61–116. Springer International Publishing, 2014.
[2] J.-F. Cai, E. J. Candès, and Z. Shen. A singular value thresholding algorithm for matrix completion. *SIAM Journal on Optimization*, 20(4):1956–1982, 2010.
[3] A. Daniilidis, N. Hadjisavvas, and J.-E. Martinez-Legaz. An appropriate subdifferential for quasiconvex functions. *SIAM Journal on Optimization*, 12(2):407–420, 2002.
[4] Z. Lin, M. Chen, and Y. Ma. The augmented Lagrange multiplier method for exact recovery of corrupted low-rank matrices. Technical Report UILU-ENG-09-2215, UIUC, 2009.
[5] T.-H. Oh, Y.-W. Tai, J.-C. Bazin, H. Kim, and I. S. Kweon. Partial sum minimization of singular values in robust PCA: Algorithm and applications. *IEEE Transactions on Pattern Analysis and Machine Intelligence (TPAMI)*, 2016.
[6] B. Wu, C. Ding, D. Sun, and K.-C. Toh. On the moreau–yosida regularization of the vector k-norm related functions. *SIAM Journal on Optimization*, 24(2):766–794, 2014.